\documentclass[sigconf,screen,nonacm]{acmart}

\usepackage{multirow}
\usepackage{float}
\usepackage{subcaption}

\AtBeginDocument{%
  }

\begin{document}

\title{FluentAvatar: Flicker-Free Talking-Head Animation via Phoneme-Guided Autoregressive Modeling}

\author{Yuchen Deng}
\authornote{These authors contributed equally to this work.}
\affiliation{%
  \institution{Shenzhen International Graduate School, Tsinghua University}
  \city{Shenzhen}
  \country{China}}
\affiliation{%
  \institution{Pengcheng Laboratory}
  \city{Shenzhen}
  \country{China}}
\email{dyc23@mails.tsinghua.edu.cn}

\author{Xiuyang Wu}
\authornotemark[1]
\affiliation{%
  \institution{Shenzhen International Graduate School, Tsinghua University}
  \city{Shenzhen}
  \country{China}}
\email{xiuyangwu23@gmail.com}

\author{Hai-Tao Zheng}
\affiliation{%
  \institution{Shenzhen International Graduate School, Tsinghua University}
  \city{Shenzhen}
  \country{China}}
\affiliation{%
  \institution{Pengcheng Laboratory}
  \city{Shenzhen}
  \country{China}}
\email{zheng.haitao@sz.tsinghua.edu.cn}

\author{Suiyang Zhang}
\affiliation{%
  \institution{Shenzhen International Graduate School, Tsinghua University}
  \city{Shenzhen}
  \country{China}}
\email{suiyangzh@gmail.com}

\author{Yi He}
\affiliation{%
  \institution{Shenzhen International Graduate School, Tsinghua University}
  \city{Shenzhen}
  \country{China}}
\email{heey754189157@gmail.com}

\author{Yuxing Han}
\authornote{Corresponding author.}
\affiliation{%
  \institution{Shenzhen International Graduate School, Tsinghua University}
  \city{Shenzhen}
  \country{China}}
\email{yuxinghan@sz.tsinghua.edu.cn}

\renewcommand{\shortauthors}{Deng et al.}

\begin{abstract}
Current talking-head generation has gradually shifted from GAN-based methods to diffusion-based paradigms, achieving remarkable progress in visual fidelity and temporal consistency. However, inter-frame flicker remains prevalent in existing diffusion-based methods. An important reason is that denoising trajectory variation induced by stochastic initialization leaves residual inter-frame inconsistencies, which manifest as short-term, abrupt visual fluctuations between adjacent frames. To further verify this, we conduct a controlled study by fixing the input while varying only the random seed. The results show markedly different flicker patterns across samplings, with a mean inter-seed Pearson correlation of only $r=0.15$. This motivates us to explore autoregressive generation, which models frames sequentially and provides a more direct prior for temporal continuity. Based on this, we propose FluentAvatar, a two-stage autoregressive framework built on phoneme representations. First, Facial Keyframe Generation produces phoneme-aligned keyframes under a Phoneme-Frame Causal Attention Mask, and Inter-frame Interpolation synthesizes transition frames via a timestamp-aware adaptive strategy built upon selective state space modeling. Moreover, we introduce BG-Flicker, a background-isolated metric for talking-head videos that enables more reliable evaluation of inter-frame flicker. Experiments on CMLR and HDTF demonstrate that FluentAvatar achieves strong performance in visual fidelity, lip synchronization, and temporal stability, attaining the best FVD on both datasets and BG-Flicker results close to ground truth. The code, the model, and the interface will be released to facilitate further research.

\end{abstract}

\begin{teaserfigure}
  \includegraphics[width=\textwidth]{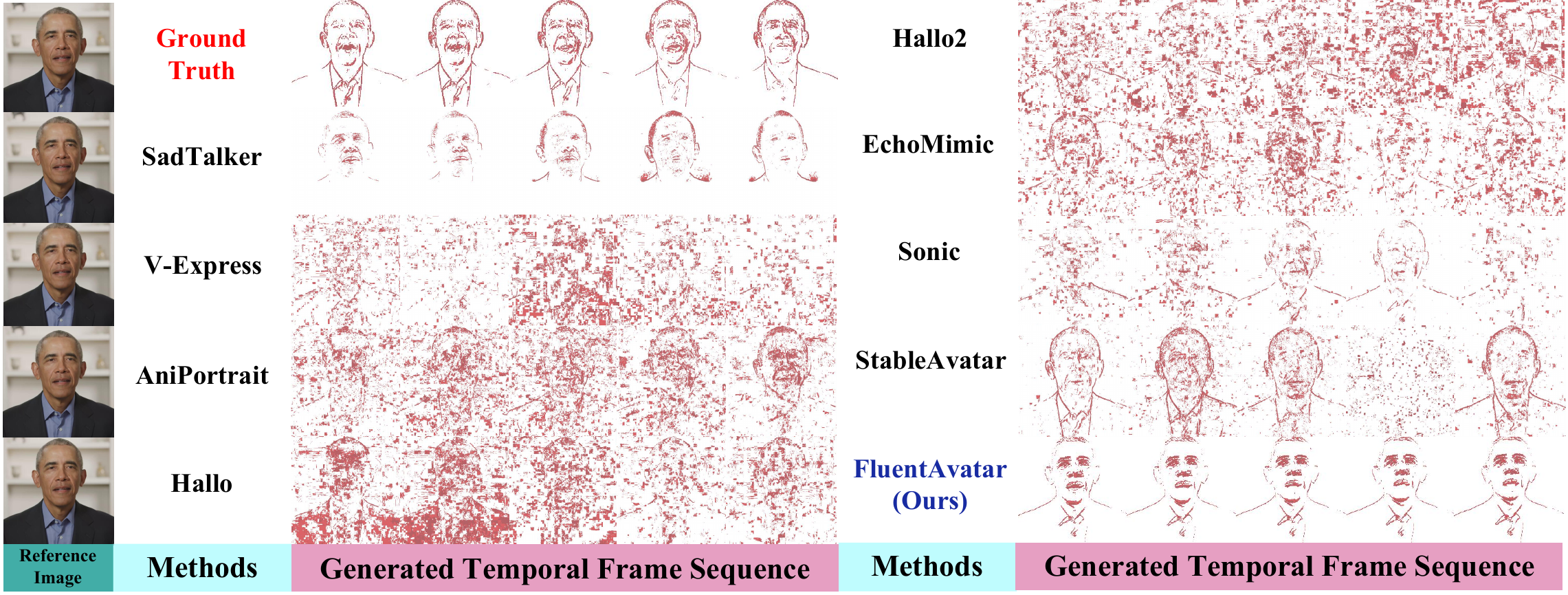}
  \caption{Inter-frame flickering across methods. Each row shows pixel-wise differences between consecutive generated frames. FluentAvatar (Ours) exhibits negligible inter-frame flicker compared to GAN and diffusion baselines.}
  \Description{Different Methods of Inter-frame Flickering.}
  \label{fig:Flickering}
\end{teaserfigure}

\maketitle

\section{Introduction}
\begin{figure}[t]
  \centering
  \setlength{\abovecaptionskip}{6pt}
  \setlength{\belowcaptionskip}{-7pt}
  \includegraphics[width=\columnwidth,keepaspectratio]{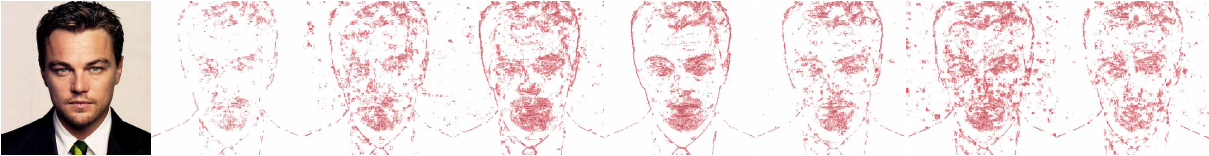}
  \caption{
    Inter-frame Flicker Visualization. Left: reference frame; subsequent panels show pixel-wise differences between consecutive frames, where scattered high-difference regions reveal flicker.
  }
  \label{fig:Frames_Diff_Heatmap}
\end{figure}

Talking-head animation~\cite{guo2024liveportrait,hu2024animate,tian2024emo,wei2024aniportrait,wang2025instructavatar} is a representative multimodal generation task that synthesizes realistic, audio-synchronized facial motion from a single static image and audio input. This technology has broad applications in video dubbing~\cite{zhang2024musetalk}, virtual avatars~\cite{zhang2024impact}, and digital entertainment~\cite{prajwal2020lip}. Despite substantial progress, a key challenge remains: \textbf{inter-frame flicker}, which manifests as short-term, abrupt visual fluctuations between adjacent frames, particularly in the background and other regions that are expected to remain relatively stable, as illustrated in Figure~\ref{fig:Frames_Diff_Heatmap}. Notably, even when individual frames exhibit high visual quality, such localized temporal irregularities can still be perceived during playback. This significantly undermines perceptual realism and limits the deployment of existing methods.

\begin{figure}[t]
  \centering
  \setlength{\abovecaptionskip}{5pt}
  \begin{minipage}[t]{0.58\columnwidth}
    \centering
    \includegraphics[width=\columnwidth]{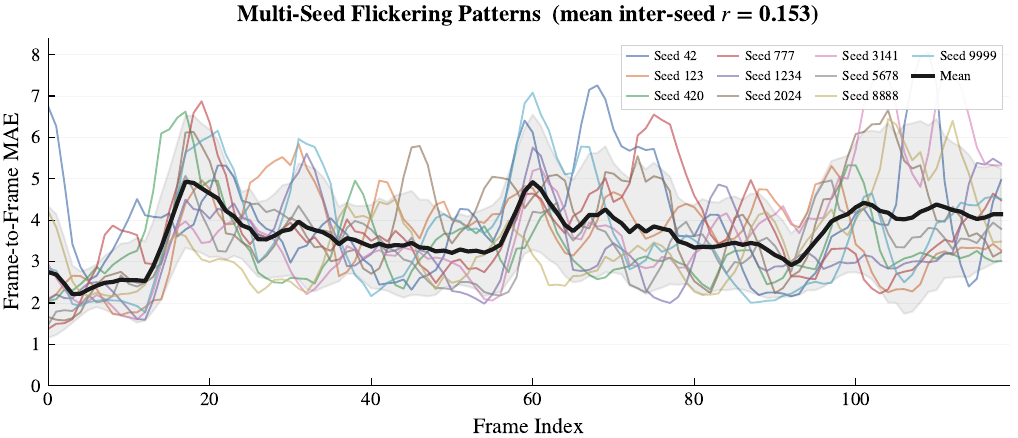}
    \centerline{\small (a) Per-frame flickering across 10 seeds}
  \end{minipage}
  \hfill
  \begin{minipage}[t]{0.38\columnwidth}
    \centering
    \includegraphics[width=\columnwidth]{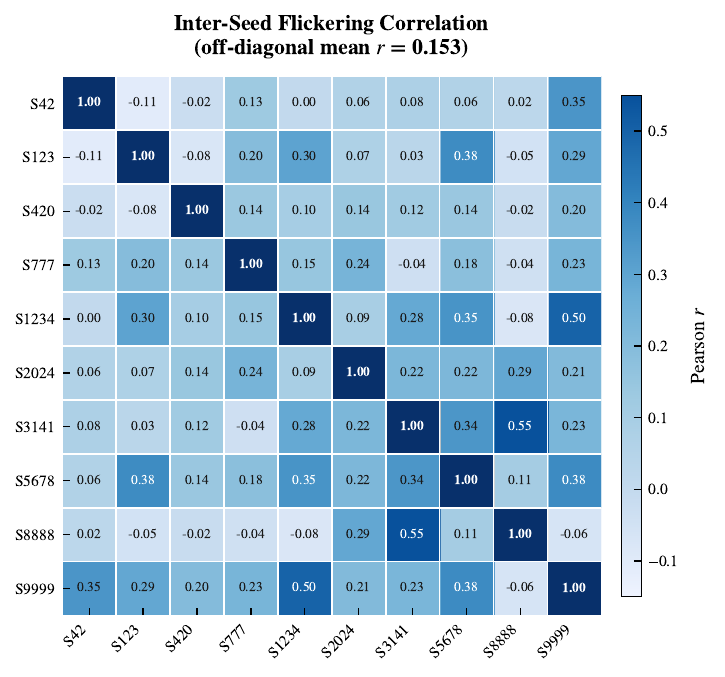}
    \centerline{\small (b) Inter-seed correlation}
  \end{minipage}
  \caption{Multi-seed flickering analysis on EchoMimic with identical inputs. (a) Per-frame MAE curves from 10 seeds show entirely different flickering patterns. (b) Inter-seed Pearson correlation (mean off-diagonal $r = 0.15$) confirms flickering is noise-dependent.}
  \label{fig:multi_seed}
  \vspace{-9pt}
\end{figure}

Current talking-head generation~\cite{tian2024emo,xu2024hallo,jiang2024loopy,lin2025omnihuman} is dominated by diffusion-based methods~\cite{ho2020denoising}, which have achieved remarkable gains in visual fidelity. Moreover, to improve temporal consistency, recent approaches have evolved from local temporal enhancement strategies, such as temporal attention~\cite{tian2024emo,xu2024hallo,chen2025echomimic} and audio adaptation~\cite{tu2025stableavatar}, to stronger architectures and generation paradigms, including Diffusion Transformers~\cite{ji2024sonic,gan2025omniavatar} and autoregressive diffusion~\cite{huang2025self,guo2025arig,huang2025live}. Overall, these advances have substantially alleviated blur, ghosting, and identity drift. However, inter-frame flicker, as a finer-grained artifact at the level of visual details, remains insufficiently addressed in existing diffusion-based methods.
From a mechanistic perspective, despite conditioning inputs, shared noise~\cite{li2024latentsync}, and temporal attention~\cite{ji2024sonic} for cross-frame coupling, stochastic initialization can still induce divergent denoising trajectories, leaving residual inconsistencies between adjacent frames. To examine this effect, we conduct a controlled study on EchoMimic~\cite{chen2025echomimic}, a representative diffusion baseline equipped with temporal attention. Keeping the reference image and audio fixed, we generate 10 videos by varying only the random seed. As illustrated in Figure~\ref{fig:multi_seed}, frame-wise flicker patterns differ markedly across different samplings. Pearson correlation analysis on the frame-wise MAE curves further quantifies this effect: the mean off-diagonal correlation is only $r=0.15$, with values ranging from $-0.11$ to $0.55$. These results indicate that denoising trajectory variation induced by stochastic initialization is an important source of inter-frame flicker in diffusion-based talking-head generation.

Beyond the diffusion paradigm, recent progress in video generation has increasingly highlighted sequential generation as a promising modeling direction. Works such as VideoGPT~\cite{yan2021videogpt}, MAGVIT~\cite{yu2023magvit}, and Open-MAGVIT2~\cite{luo2024open} show that sequential modeling over discrete visual tokens~\cite{tian2024visual} can be highly effective in general video generation. More specifically, autoregressive generation models frames sequentially, providing a more direct prior for cross-frame dependency and temporal continuity than frame-wise noise-driven synthesis. Motivated by these, we turn to the autoregressive paradigm to explore a generation path more suitable for temporal modeling in talking-head animation.

We introduce \textbf{FluentAvatar}, a two-stage autoregressive framework built on phoneme representations for generating natural, stable, and controllable talking-head videos from a single reference image and audio/text input. Phonemes, as the minimal units of speech articulation, naturally capture key states in lip shape and facial motion evolution. As shown in Figure~\ref{fig:framework}, FluentAvatar takes phoneme sequences and the discrete visual representation of the reference image as unified inputs and synthesizes videos in two stages. In the first stage, Facial Keyframe Generation (FKG) produces phoneme-aligned keyframes under a Phoneme-Frame Causal Attention Mask, explicitly modeling speech-driven key visual states. In the second stage, Inter-frame Interpolation conditions on these keyframes and employs a timestamp-aware adaptive interpolation strategy built upon selective state space modeling to generate transition frames, yielding continuous and natural talking-head videos. By decoupling key-state modeling from local motion transition generation, this two-stage design improves temporal coherence, mitigates error accumulation in long-sequence autoregressive generation, and supports fine-grained, segment-level control.

Additionally, existing metrics~\cite{huang2024vbench} are designed for general videos and measure pixel variation over entire frames. However, this paradigm is not suited to talking-head videos, where head motion, facial expressions, and lip articulation introduce substantial pixel changes unrelated to flicker. Therefore, we propose \textbf{BG-Flicker}, a background-isolated metric that restricts evaluation to relatively stable regions, where pixel variations can be more reliably attributed to flickering artifacts. Experiments on Chinese (CMLR~\cite{zhao2019cascade}) and English (HDTF~\cite{zhang2021flow}) benchmarks demonstrate that FluentAvatar achieves strong performance in visual fidelity, lip synchronization, and temporal stability, while attaining the best FVD on both datasets (CMLR: 503.29; HDTF: 260.65). Under BG-Flicker, FluentAvatar substantially reduces flicker error relative to representative diffusion baselines, with scores matching ground truth on CMLR (0.9981) and remaining near-identical on HDTF (0.9988 vs.\ 0.9991). 

In summary, our main contributions are listed as follows:
\begin{itemize}
\item Mechanism analysis and controlled experiments show that denoising trajectory variation induced by stochastic initialization is an important source of inter-frame flicker in diffusion-based generation. Based on this, we propose \textbf{FluentAvatar}, a phoneme-based autoregressive framework for improving temporal stability in talking-head videos.

\item We design a two-stage pipeline: Facial Keyframe Generation produces phoneme-aligned keyframes under a Phoneme-Frame Causal Attention Mask, and Inter-frame Interpolation synthesizes transition frames via a timestamp-aware adaptive strategy built upon selective state space modeling. This design mitigates error accumulation and enables fine-grained control.

\item We introduce BG-Flicker, a background-isolated metric that adapts general video flicker evaluation to the talking-head setting by reducing interference from legitimate motion.

\item Extensive experiments on CMLR and HDTF demonstrate that FluentAvatar achieves superior performance in visual fidelity, lip synchronization, and temporal stability, including the best FVD on both benchmarks and BG-Flicker errors close to ground truth.
\end{itemize}

\section{Related work}
\subsection{Talking Head Generation}
Audio-driven talking-head generation has emerged as a key research topic in multimodal content generation, demonstrating significant practical value in applications such as video dubbing and virtual avatars. Prevailing approaches can be broadly categorized into two classes: GAN-based methods~\cite{zhou2019talking,zhou2021pose,meshry2021learned,das2020speech,chen2019hierarchical,zhang2023sadtalker} and diffusion-based methods~\cite{wang2024v,ji2024sonic,lin2025omnihuman,li2024latentsync,jiang2024loopy,xu2024vasa}.

\paragraph{GAN-based methods.} GAN-based methods are widely recognized for their computational efficiency and rapid inference. However, early approaches struggle with maintaining identity consistency and accurate lip synchronization. To address these issues, methods such as SadTalker~\cite{zhang2023sadtalker} and FaceVid2Vid~\cite{wang2021one} adopt multi-stage inference pipelines that decouple audio-to-motion modeling from motion-to-video synthesis. While this improves generation quality, the fully decoupled design often limits global dynamics, causing lip movements to dominate and leaving the head, background, and other regions largely static, thereby weakening motion naturalness and temporal coherence.

\paragraph{Diffusion-based methods.} Early diffusion-based approaches typically integrate a ReferenceNet, temporal modeling layers, and audio attention modules into a unified U-Net framework~\cite{tian2024emo,xu2024hallo,cui2024hallo2,lin2025omnihuman,jiang2024loopy,xu2024vasa}. V-Express~\cite{wang2024v} employs conditional dropout or cyclic prediction to balance audio and visual conditions. EchoMimic~\cite{chen2025echomimic} uses gesture sequences to improve the quality of video generation. Recent works have gradually shifted from U-Net to Diffusion Transformer (DiT)~\cite{peebles2023scalable} architectures for stronger scalability. Sonic~\cite{ji2024sonic} emphasizes global audio perception to improve motion naturalness. StableAvatar~\cite{tu2025stableavatar} further enhances temporal stability in long-video generation through the introduction of an audio adapter. Recent studies have also investigated autoregressive diffusion~\cite{huang2025self,guo2025arig,huang2025live,li2025ditto} for real-time or streaming generation. Although these methods improve visual fidelity and temporal coherence, they do not fully eliminate the temporal instability inherent to iterative denoising, still suffering from inter-frame flicker. 

\subsection{Visual Generation Based on LLMs}
Large language models (LLMs) ~\cite{achiam2023gpt,touvron2023llama,liang2024survey} have extended to the domain of visual generation, offering superior scalability over multi-step denoising.
Masked language models such as MaskGIT~\cite{chang2022maskgit} enable efficient parallel sampling, while autoregressive approaches~\cite{sun2024autoregressive,Tian2024} have demonstrated competitive image quality. This paradigm has been further extended to video generation~\cite{kondratyuk2023videopoet,hong2022cogvideo,xie2024show,wang2024emu3}, with multimodal architectures for text-to-video synthesis.

Recently, a few studies have explored autoregressive architectures for talking-head generation. ARTalk~\cite{chu2025artalk} applies autoregressive modeling to 3D head animation rather than our 2D portrait setting. Teller~\cite{zhen2025teller} proposes a streaming autoregressive approach for 2D portraits, but the lack of publicly available code and models prevents direct comparison. Both methods typically predict visual tokens step by step under the full audio condition, making them prone to error accumulation and limiting fine-grained temporal control. In contrast, FluentAvatar introduces phonemes as an explicit intermediate representation and adopts a two-stage framework that decouples semantic alignment from visual dynamics, thereby alleviating error accumulation through parallel interpolation and enabling fine-grained, segment-level editing.

\section{Method}
\label{headings}
\subsection{Why Diffusion Models Flicker}
\label{sec:flickering_analysis}
As illustrated in Figure~\ref{fig:Frames_Diff_Heatmap}, diffusion-based talking-head methods exhibit pervasive inter-frame flickering. We provide both theoretical and empirical analysis to identify the root cause.

\paragraph{Theoretical Analysis.}
Consider the standard diffusion reverse process for generating frame $t$:
\begin{equation}
\mathbf{x}_{T}^{(t)} \sim \mathcal{N}(\mathbf{0}, \mathbf{I}), \quad \hat{\mathbf{x}}^{(t)}_0 = f_\theta(\mathbf{x}_{T}^{(t)}, \mathbf{c}^{(t)})
\end{equation}
where $\mathbf{x}_T^{(t)}$ is the initial noise for frame $t$, $\mathbf{c}^{(t)}$ is the conditioning input, and $f_\theta(\cdot)$ denotes the denoising mapping. Since each frame is initialized from independently sampled Gaussian noise, the covariance between adjacent frames' initial latents is zero:
\begin{equation}
\text{Cov}(\mathbf{x}_T^{(t)}, \mathbf{x}_T^{(t+1)}) = \mathbf{0}
\end{equation}
This means adjacent frames share no statistical dependency at initialization. Recent extensions mitigate this issue through shared noise~\cite{li2024latentsync} and cross-frame temporal coupling~\cite{chen2025echomimic}, such as temporal attention, but cannot eliminate the resulting stochastic mismatch, as cross-frame alignment remains approximate~\cite{ge2023preserve}. Under the highly nonlinear denoising mapping $f_\theta$, this mismatch can still drive adjacent frames along different generation trajectories, producing residual discrepancies that manifest as inter-frame flickering. As a result, such flickering is better understood as a structural phenomenon of diffusion-based video generation rather than merely incidental noise.

\paragraph{Empirical Verification.}
To verify this analysis, we conduct controlled experiments on EchoMimic~\cite{chen2025echomimic}, a representative diffusion baseline with temporal attention. Using the identical inputs (same reference image and audio), we generate 10 videos with different random seeds and analyze the resulting flickering patterns.

Finding 1: Noise divergence does not leak spatially. We compute pixel-wise divergence maps between adjacent frames for both the initial noise and the final output, and then measure their spatial correlation. Across five seeds, the mean Pearson correlation is only $r = 0.033$ (Table~\ref{tab:noise_corr}), indicating that, after temporal attention and conditional guidance, the spatial pattern of inter-frame discrepancies in the final output no longer directly preserves that of the initial noise. Flickering does not arise from direct spatial leakage of independent noise.

\begin{table}[H]
  \centering
  \setlength{\abovecaptionskip}{4pt}
  \setlength{\belowcaptionskip}{-5pt}
  \caption{Spatial correlation between initial noise divergence and output flickering in EchoMimic. Near-zero Pearson $r$ indicates no direct spatial leakage.}
  \label{tab:noise_corr}
  \small
  \begin{tabular}{cccc}
    \toprule
    Seed & Pearson $r$ & Noise Div. & Flicker \\
    \midrule
    42   & 0.035 & 99.6 & 4.13 \\
    123  & 0.042 & 99.7 & 3.75 \\
    420  & 0.023 & 99.6 & 3.37 \\
    777  & 0.029 & 99.6 & 3.77 \\
    1234 & 0.034 & 99.7 & 3.34 \\
    \midrule
    Mean & 0.033 $\pm$ 0.006 & 99.6 & 3.67 \\
    \bottomrule
  \end{tabular}
\end{table}

Finding 2: Flickering is seed-dependent. Although the spatial pattern of the initial noise divergence is not directly preserved in the final output, flickering still persists and remains highly sensitive to the random seed. As shown in Figure~\ref{fig:multi_seed}(a), the per-frame MAE curves obtained from 10 different seeds exhibit markedly different temporal fluctuation patterns under identical inputs. To quantify this effect, we compute the Pearson correlation between the per-frame MAE curves of every pair of seeds. The resulting inter-seed correlation matrix in Figure~\ref{fig:multi_seed}(b) shows a mean off-diagonal correlation of only $r = 0.15$, with values ranging from $-0.11$ to $0.55$. This indicates that changing only the noise initialization leads to substantially different flickering trajectories under otherwise identical conditions.

\paragraph{Implications.}
Together, these two findings reveal a key mechanism underlying flickering in diffusion-based generation: stochastic variation introduced at noise initialization drives frames along different denoising trajectories, and temporal attention, while able to compress the resulting inter-frame discrepancies, cannot fully remove them. The residual inconsistency is further reshaped by the nonlinear denoising process and manifested as visible inter-frame flicker. Because noise initialization remains a core component of diffusion generation, the resulting temporal instability continues to pose a persistent challenge within this paradigm. This motivates exploring alternative generation paradigms that do not rely on frame-wise noise initialization.

\paragraph{Autoregressive Alternative.}
Autoregressive models generate video frames as a unified token sequence, where each token is conditioned on all preceding tokens:
\begin{equation}
P(X) = \prod_{i=1}^{N} P(x_i \mid x_{<i}, c)
\end{equation}
For any token $x_j^{(t)}$ in frame $t$:
\begin{equation}
P(x_j^{(t)} \mid x_1^{(1)}, \ldots, x_K^{(t-1)}, x_1^{(t)}, \ldots, x_{j-1}^{(t)}, c)
\end{equation}
There is no per-frame noise initialization and no parallel denoising from independent starting points. Each frame is generated sequentially, directly conditioned on the complete results of all previously generated frames. This sequential dependency makes temporal coherence an intrinsic property of the generation process rather than a post-hoc constraint, thereby structurally avoiding the primary source of flickering identified above.

\begin{figure*}[t]
  \vspace{-10pt}
  \centering
  \setlength{\abovecaptionskip}{5pt}
  \includegraphics[width=\textwidth,keepaspectratio]{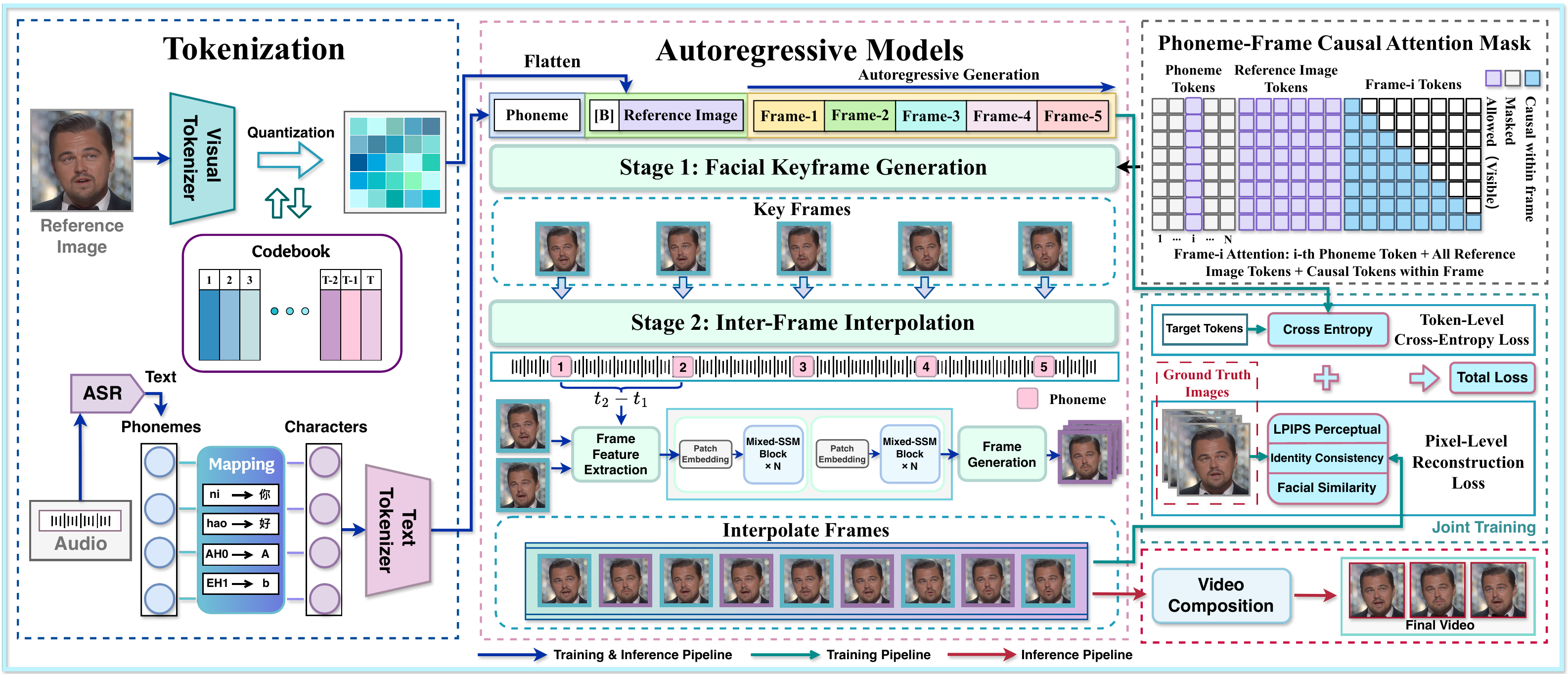}
  \caption{
  The overall framework of FluentAvatar. The pipeline first normalizes text/audio into a phoneme token sequence via a many-to-one mapping, and tokenizes the reference image into visual tokens. Next, a two-stage autoregressive generator performs Facial Keyframe Generation under a Phoneme-Frame Causal Attention Mask, and inserts intermediate frames using a timestamp-aware module that interleaves keyframes for linear-time global context. Finally, the decoder reconstructs RGB frames to animate the character.}
  \label{fig:framework}
  \vspace{-5pt}
\end{figure*}

\subsection{BG-Flicker: Background-Isolated Flickering Metric}
\label{sec:bg_flicker_method}
Existing temporal consistency metrics~\cite{huang2024vbench} compute pixel-level variations across entire frames. However, in talking-head videos, lip articulation, facial expressions, and head movement introduce substantial inter-frame pixel changes that reflect intended motion rather than flickering artifacts, which can confound measurements. To address this, we propose BG-Flicker, a task-specific metric that evaluates temporal stability only in background regions, where pixel variation is less affected by semantic facial motion and thus serves as a cleaner proxy for flickering.

Specifically, for each frame we apply DeepLabV3~\cite{chen2017rethinking} with a ResNet-101 backbone to obtain a semantic segmentation map, and define the background mask as all pixels not classified as person. To avoid noise from segmentation boundary jitter, we compute the intersection of adjacent frames' background masks. For each consecutive frame pair $(t, t+1)$, the background MAE is:
\begin{equation}
\text{MAE}(t) = \frac{1}{|\mathcal{B}_t|} \sum_{p \in \mathcal{B}_t} \left| I^{(t)}(p) - I^{(t+1)}(p) \right|
\end{equation}
where $\mathcal{B}_t = \text{bg\_mask}^{(t)} \cap \text{bg\_mask}^{(t+1)}$ denotes the intersection of background regions and $I^{(t)}(p)$ is the pixel value at position $p$ in frame $t$. The final BG-Flicker score aggregates over all frame pairs and normalizes to $[0, 1]$:
\begin{equation}
\text{BG-Flicker} = 1 - \frac{1}{(T-1) \cdot 255} \sum_{t=1}^{T-1} \text{MAE}(t)
\end{equation}
A score closer to 1 indicates a more temporally stable background (less flickering). By excluding the foreground person region, BG-Flicker better distinguishes inter-frame flicker from intended facial motion, providing a more faithful evaluation of temporal consistency in talking-head videos.

\subsection{Model Design}

\paragraph{Overview.}
The overall framework of FluentAvatar is depicted in Figure~\ref{fig:framework}. Given a single reference image and either audio or text input, FluentAvatar generates temporally coherent talking-head videos through a two-stage autoregressive pipeline built on phoneme representations. The pipeline consists of three components: (1) a \textit{tokenization} module that converts audio or text into a phoneme sequence and the reference image into discrete visual tokens; (2) a \textit{two-stage autoregressive generator} comprising Facial Keyframe Generation (FKG) and Inter-frame Interpolation; and (3) a \textit{decoder} that reconstructs RGB frames from the generated tokens.

The key design principle is that phonemes, as the minimal units of speech articulation, serve as natural anchors for temporal modeling. Specifically, each phoneme anchors a keyframe, while smooth motion is synthesized by interpolating between adjacent keyframes. This two-stage design decouples semantic alignment from visual dynamics, enabling fine-grained control while mitigating error accumulation through keyframe anchoring. Notably, FluentAvatar's temporal consistency arises not from any single component but from the synergy of the entire design: the autoregressive paradigm avoids per-frame noise initialization (Section~\ref{sec:flickering_analysis}); the Phoneme-Frame Causal Attention Mask limits cross-frame information leakage; the timestamp-aware interpolation ensures rhythm-aligned motion with improved temporal smoothness; and the two-stage keyframe anchoring mitigates error accumulation. Together, these components make temporal coherence a structural property of the pipeline rather than a post-hoc correction.

\paragraph{Tokenization.} FluentAvatar supports both text-image and audio-image input modalities.
\textit{(1) Phoneme Extraction.} For text input, the text is first converted into a phoneme sequence via a deterministic many-to-one mapping, then tokenized into discrete tokens. For audio input, we employ ASR tools to extract phoneme-level alignments with timestamps, which are tokenized into discrete phoneme tokens.
Additionally, for coarticulation patterns with strong visual manifestation, we define context-dependent phoneme mappings before tokenization, so that these local articulatory interactions can be explicitly encoded in the subsequent generation process. In both cases, phonemes serve as the unified intermediate representation, ensuring consistent phoneme-to-visual alignment regardless of input modality.
\textit{(2) Image Tokenization.} We adopt a pre-trained vision foundation model (Open-MAGVIT2~\cite{luo2024open}) to quantize the reference image into discrete visual tokens. To capture fine-grained facial details, we employ MMPose for facial landmark detection and adjust the input image's aspect ratio accordingly.

\begin{table*}[t]
\centering
\caption{Quantitative comparison on CMLR and HDTF datasets. SadTalker is GAN-based, the other baselines are diffusion models, and FluentAvatar (ours) is autoregressive.}
\label{tab:quantitative_results}
\vspace{-6pt}
\resizebox{\textwidth}{!}{%
  \renewcommand{\arraystretch}{1.15}%
  \begin{tabular}{l cccccccc cccccccc}
  \toprule
  \multirow{2}{*}{\textbf{Method}} &
  \multicolumn{8}{c}{\textbf{CMLR}} &
  \multicolumn{8}{c}{\textbf{HDTF}} \\
  \cmidrule(lr){2-9} \cmidrule(lr){10-17}
  & FID$\downarrow$ & FVD$\downarrow$ & LPIPS$\downarrow$ & PSNR$\uparrow$ & SSIM$\uparrow$ & CSIM$\uparrow$ & Sync-C$\uparrow$ & Sync-D$\downarrow$
  & FID$\downarrow$ & FVD$\downarrow$ & LPIPS$\downarrow$ & PSNR$\uparrow$ & SSIM$\uparrow$ & CSIM$\uparrow$ & Sync-C$\uparrow$ & Sync-D$\downarrow$ \\
  \midrule
  SadTalker        & 30.96 & 618.74  & \underline{0.42} & 14.50 & 0.48 & \underline{0.79} & 0.91 & 14.29 & 37.20 & 316.78 & 0.26 & 18.85 & 0.74 & 0.82 & 5.72 & 11.83 \\
  V-Express        & 32.54 & 821.95  & 0.46 & 12.99 & 0.39 & 0.70 & 0.83 & 13.71 & 40.13 & 779.19 & 0.27 & 18.29 & 0.72 & 0.88 & 5.29 & 12.79 \\
  AniPortrait      & 32.46 & 711.30  & 0.47 & 14.71 & \textbf{0.49} & 0.69 & 0.89 & 14.99 & 40.19 & 614.01 & 0.28 & 18.52 & 0.72 & 0.88 & 5.61 & \underline{11.30} \\
  Hallo            & \underline{30.84} & 630.53  & 0.43 & 14.62 & \textbf{0.49} & 0.71 & \underline{1.20} & 14.99 & 44.48 & 520.72 & \underline{0.25} & \underline{18.94} & \textbf{0.75} & 0.88 & 5.05 & 11.62 \\
  Hallo2           & 35.10 & \underline{570.12}  & 0.44 & 14.59 & \textbf{0.49} & 0.75 & 0.80 & 14.67 & 42.77 & 340.83 & \underline{0.25} & \textbf{19.04} & \textbf{0.75} & 0.89 & \underline{6.37} & 11.91 \\
  EchoMimic        & 33.09 & 1225.10 & 0.42 & \underline{14.87} & 0.48 & 0.69 & 0.89 & 13.97 & 38.11 & \underline{301.33} & 0.29 & 18.11 & 0.73 & 0.85 & 2.63 & 12.88 \\
  Sonic            & 31.58 & 953.30  & 0.43 & 14.39 & 0.47 & 0.68 & \textbf{2.25} & \underline{12.84} & \underline{32.89} & 379.13 & 0.26 & 18.55 & 0.74 & \underline{0.90} & \textbf{7.65} & 12.70 \\
  StableAvatar     & 39.38 & 741.36  & 0.46 & 14.41 & \textbf{0.49} & \underline{0.79} & 0.92 & 14.83 & 35.39 & 700.82 & 0.28 & 18.41 & 0.72 & 0.85 & 5.34 & 12.56 \\
  \textbf{FluentAvatar} & \textbf{25.19} & \textbf{503.29}  & \textbf{0.40} & \textbf{15.07} & \textbf{0.49} & \textbf{0.85} & 0.94 & \textbf{10.90} & \textbf{29.72} & \textbf{260.65} & \textbf{0.24} & 18.69 & 0.73 & \textbf{0.95} & 5.57 & \textbf{10.39} \\
  \bottomrule
  \end{tabular}%
}
\end{table*}

\paragraph{Two-Stage Autoregressive Generation.} 
\textbf{Stage 1: Facial Keyframe Generation (FKG).} 
The FKG module generates $T_s$ keyframes aligned with the order of the input phonemes. The input sequence is structured as: \{\texttt{\{Phoneme\} [B] \{Frame$_1$\}, ..., \{Frame$_{T_s}$\}}\}, where each keyframe is paired with its corresponding phoneme.
The core innovation of this stage is the \textbf{Phoneme-Frame Causal Attention Mask}, which enforces a strict one-to-one mapping between phonemes and keyframes. Specifically, when generating each keyframe, the model attends only to its paired phoneme and the reference image, with cross-frame attention masked to prevent information leakage. This design ensures precise phoneme-to-visual alignment: the model learns to produce the mouth shape, facial expression, and head pose that correspond to each specific phoneme, rather than blending information across the entire sequence. Unlike prior two-stage designs~\cite{harvey2022flexible,wei2024aniportrait} that uniformly sample keyframes in time, our keyframes are explicitly anchored to phoneme units, making the generated skeleton semantically meaningful.

\textbf{Stage 2: Inter-frame Interpolation.}
Given the phoneme-aligned keyframes, the interpolation module synthesizes transition frames to produce temporally smooth video. The core innovation is the \textbf{timestamp-aware adaptive strategy} built upon selective state space modeling~\cite{zhang2024vfimamba}.
This design addresses a key challenge: the temporal intervals between adjacent phonemes are variable (some phonemes are held longer than others), so the number and timing of transition frames must adapt to the phoneme rhythm. Guided by phoneme-timestamp pairs, the module dynamically determines the number of intermediate frames to insert between each keyframe pair. At each interpolation step, adjacent keyframes are encoded into interleaved token sequences and processed via state space modeling, enabling efficient global context aggregation with linear complexity. This ensures that the synthesized motion intensity matches the actual duration of each phoneme interval, maintaining natural speech rhythm and stable output frame rates.

Since both stages employ autoregressive generation where each output token is conditioned on all preceding tokens, temporal consistency is structurally maintained throughout the pipeline. Furthermore, interpolations between different keyframe pairs can be performed in parallel, significantly improving inference efficiency.

\subsection{Training Strategy}

We decouple the training into two phases to address semantic accuracy and visual refinement separately.

\paragraph{Phase 1: Semantic Learning.}
The FKG module is first trained with a token-level reconstruction loss to learn phoneme-conditioned facial generation:
\begin{equation}
\mathcal{L}_{\mathit{recon}} = - \sum_{i} \log P\left( v_i^{\mathit{real}} \mid \mathbf{x} \right)
\end{equation}
where $v_i^{\mathit{real}}$ is the ground-truth token at position $i$. This phase focuses on learning the abstract mapping from phonemes to facial structure without pixel-level supervision.

\paragraph{Phase 2: Visual Refinement.}
Once semantic learning converges, we introduce three pixel-space losses to enhance visual quality. The LPIPS perceptual loss~\cite{zhang2018unreasonable} improves fine-grained visual fidelity:
\begin{equation}
\mathcal{L}_{\mathit{lpips}} = \sum_{l} w_l \cdot \frac{1}{H_l W_l} \sum_{h,w} \left\| F_l(I_{\text{gen}})_{h,w} - F_l(I_{\text{real}})_{h,w} \right\|_2^2
\end{equation}
where $F_l(\cdot)$ denotes the feature map extracted from the $l$-th layer of the perceptual network, and $H_l$, $W_l$, and $w_l$ denote the corresponding spatial dimensions and layer weight. The identity consistency loss preserves the character's identity across frames:
\begin{equation}
\mathcal{L}_{Id} = \frac{1}{N} \sum_{i=1}^{N} \left(1 - \cos(f^{i}_{gen}, f^{i}_{real}) \right) \cdot w_{id}
\end{equation}
where $f^{i}_{gen}$ and $f^{i}_{real}$ denote the identity embeddings of the $i$-th generated and real frame. The facial similarity loss further enforces appearance fidelity in the FaceNet512 embedding space:
\begin{equation}
\mathcal{L}_{FS} = \frac{1}{N} \sum_{i=1}^{N} 0.5 \cdot d_{cos}(f^{i}_{gen}, f^{i}_{real}) \cdot w_{fs}
\end{equation}
where $d_{cos}(\cdot,\cdot)$ denotes cosine distance in the FaceNet512 embedding space. The overall training objective combines all terms:
\begin{equation}
\mathcal{L}_{\mathit{total}} = \lambda_{1} \cdot \mathcal{L}_{\mathit{recon}} 
+ \lambda_{2} \cdot \mathcal{L}_{\mathit{lpips}} 
+ \lambda_{3} \cdot \mathcal{L}_{\mathit{Id}} 
+ \lambda_{4} \cdot \mathcal{L}_{\mathit{FS}}
\end{equation}
This phased strategy avoids the instability of jointly optimizing multiple objectives from scratch, allowing the model to first establish robust semantic alignment before refining pixel-level quality.

\section{Experiments}
\label{others}
\subsection{Experimental Setup}
\paragraph{Training Details.} We train FluentAvatar on a mixed dataset that combines the super-resolved Chinese CMLR dataset and the original English HDTF dataset, with a standard 95:5 train-test split applied to each benchmark before mixing. The training is conducted for a total of 10,000 steps on 2 NVIDIA L20 GPUs. The model architecture consists of 48 layers with a hidden size of 3072 and 48 attention heads. At the core of training, we introduce a custom Phoneme-Frame Causal Attention Mask and utilize a composite loss function comprising a reconstruction loss (weight 1.0), a perceptual loss via LPIPS (weight 8.0), an identity preservation loss (weight 10.0), and a facial similarity loss (weight 10.0) to fine-tune the pre-trained model weights. For optimization, we employ the Adam optimizer with a learning rate of $2 \times 10^{-4}$, complemented by a cosine annealing schedule. We enable 16-bit mixed-precision training with activation checkpointing, accelerated by the DeepSpeed ZeRO-2 framework for memory efficiency. The complete training procedure is detailed in Appendix~\ref{appendix:training_process}.

\paragraph{Evaluation Metrics.} We evaluate the models with eight metrics. For perceptual quality and identity preservation, FID, FVD, LPIPS, and CSIM are computed in deep feature space, where lower FID/FVD/LPIPS and higher CSIM are better. For fidelity, PSNR and SSIM assess reconstruction accuracy with respect to the ground-truth frames in pixel space (higher is better). For lip synchronization, we adopt the SyncNet-based evaluation protocol: Sync-C measures audio-visual synchronization confidence, where higher is better, while Sync-D measures the distance between audio and visual embeddings, where lower is better. Additionally, we evaluate inter-frame flicker using our proposed BG-Flicker metric (Section~\ref{sec:bg_flicker_method}).

\paragraph{Compared Baselines.} We compare FluentAvatar with state-of-the-art audio-driven talking-head methods spanning both paradigms. For GAN-based models, we consider SadTalker~\cite{zhang2023sadtalker}. Diffusion-based baselines include V-Express~\cite{wang2024v}, AniPortrait~\cite{wei2024aniportrait}, Hallo~\cite{xu2024hallo}, Hallo2~\cite{cui2024hallo2}, EchoMimic~\cite{chen2025echomimic}, Sonic~\cite{ji2024sonic}, and StableAvatar~\cite{tu2025stableavatar}.

\subsection{Quantitative Evaluation}

\paragraph{Comparison on CMLR and HDTF} As shown in Table~\ref{tab:quantitative_results}, FluentAvatar consistently achieves state-of-the-art performance on both the Chinese CMLR and English HDTF datasets. Several results are worth highlighting in connection with our method design. First, FluentAvatar achieves the best FVD on both benchmarks (503.29 on CMLR, 260.65 on HDTF), substantially outperforming the next best methods (Hallo2: 570.12, EchoMimic: 301.33). Since FVD directly measures temporal coherence across frames, this improvement provides quantitative evidence that autoregressive sequential conditioning effectively suppresses inter-frame inconsistencies. Second, FluentAvatar achieves the highest CSIM on both datasets (0.85 on CMLR, 0.95 on HDTF), reflecting the effectiveness of the identity consistency and facial similarity losses in preserving character identity across the generated sequence. Third, the strong Sync-D scores (10.90 on CMLR, 10.39 on HDTF) validate the phoneme-based conditioning design, where the one-to-one phoneme-to-keyframe mapping provides precise articulatory alignment. These results hold across both Chinese and English, demonstrating strong cross-lingual generalization.

\paragraph{Inter-frame Flickering Evaluation.} We further evaluate temporal flickering using BG-Flicker. Since videos contain background variations, we use GT background error as the reference. As shown in Table~\ref{tab:bg_flicker}, FluentAvatar's BG-Flicker error matches the ground truth exactly on CMLR ($1.00\times$) and is within $1.33\times$ on HDTF, while diffusion baselines such as Hallo2 exhibit $3.47\times$ and $3.44\times$ the GT error respectively. We note that SadTalker achieves low BG-Flicker error, because its generation only animates the facial region while keeping the background nearly static, which trivially minimizes background pixel variation but does not reflect genuine temporal modeling capability, as evidenced by its inferior FID, FVD, and CSIM scores. The per-frame Background MAE curves in Figure~\ref{fig:bg_mae} further reveal a qualitative difference: diffusion methods exhibit frequent high-amplitude spikes characteristic of stochastic denoising, whereas FluentAvatar's curve remains consistently low and smooth, closely tracking the ground truth trajectory (avg 0.44 vs.\ GT 0.52 on CMLR; 0.30 vs.\ GT 0.27 on HDTF). This indicates that autoregressive sequential conditioning improves flicker suppression at the per-frame level, rather than only on average.

\begin{table}[t]
  \centering
  \setlength{\abovecaptionskip}{4pt}
  \setlength{\belowcaptionskip}{-5pt}
  \caption{Inter-frame Flickering Evaluation. Error = $(1-\text{BG-Flicker}) \times 10^3$; Ratio = Error / GT (closer to 1 is better).}
  \label{tab:bg_flicker}
  \small
  \setlength{\tabcolsep}{4pt}
  \begin{tabular}{lcccc}
    \toprule
    \multirow{2}{*}{Method} & \multicolumn{2}{c}{CMLR} & \multicolumn{2}{c}{HDTF} \\
    \cmidrule(lr){2-3} \cmidrule(lr){4-5}
    & Error & Ratio & Error & Ratio \\
    \midrule
    GT & 1.90 & 1.00$\times$ & 0.90 & 1.00$\times$ \\
    \midrule
    SadTalker & 0.40 & 0.21$\times$ & 1.20 & 1.33$\times$ \\
    V-Express & 2.40 & 1.26$\times$ & 1.00 & 1.11$\times$ \\
    AniPortrait & 2.80 & 1.47$\times$ & 2.40 & 2.67$\times$ \\
    Hallo & 5.40 & 2.84$\times$ & 3.00 & 3.33$\times$ \\
    Hallo2 & 6.60 & 3.47$\times$ & 3.10 & 3.44$\times$ \\
    EchoMimic & 4.80 & 2.53$\times$ & 4.80 & 5.33$\times$ \\
    Sonic & 1.20 & 0.63$\times$ & 0.80 & 0.89$\times$ \\
    StableAvatar & 1.70 & 0.89$\times$ & 1.60 & 1.78$\times$ \\
    \midrule
    \textbf{FluentAvatar} & \textbf{1.90} & $\boldsymbol{1.00\times}$ & \textbf{1.20} & $\boldsymbol{1.33\times}$ \\
    \bottomrule
  \end{tabular}
  \vspace{-8pt}
\end{table}

\begin{figure}[t]
  \centering
  \setlength{\abovecaptionskip}{4pt}
  \begin{subfigure}[t]{0.48\columnwidth}
    \includegraphics[width=\linewidth]{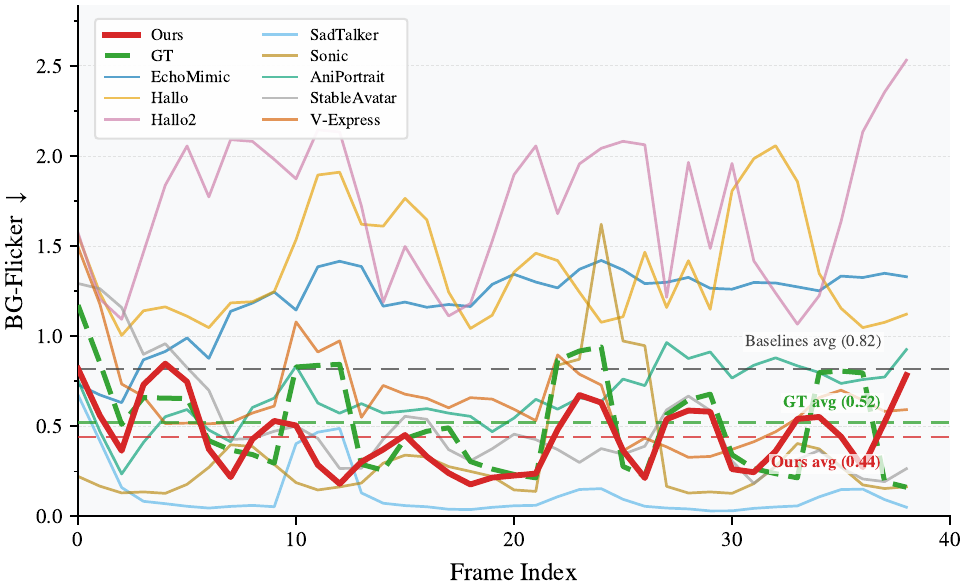}
    \caption{CMLR}
  \end{subfigure}
  \hfill
  \begin{subfigure}[t]{0.48\columnwidth}
    \includegraphics[width=\linewidth]{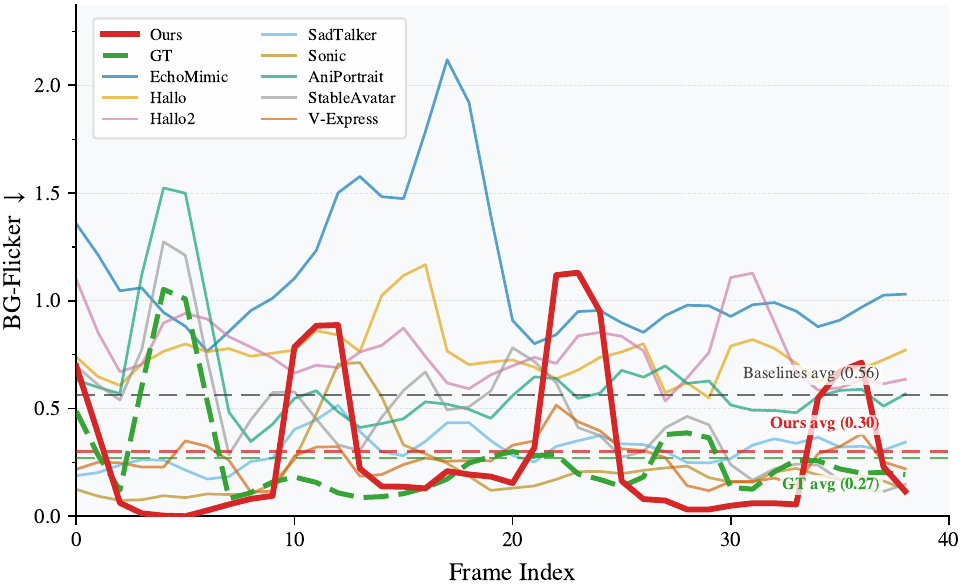}
    \caption{HDTF}
  \end{subfigure}
  \caption{Per-frame BG-Flicker on CMLR and HDTF.}
  \label{fig:bg_mae}
  \vspace{-8pt}
\end{figure}

\subsection{Qualitative Evaluation}

\begin{figure}[t!]
  \centering
  \setlength{\abovecaptionskip}{2pt}
  \setlength{\belowcaptionskip}{-10pt}
  \includegraphics[width=\columnwidth]{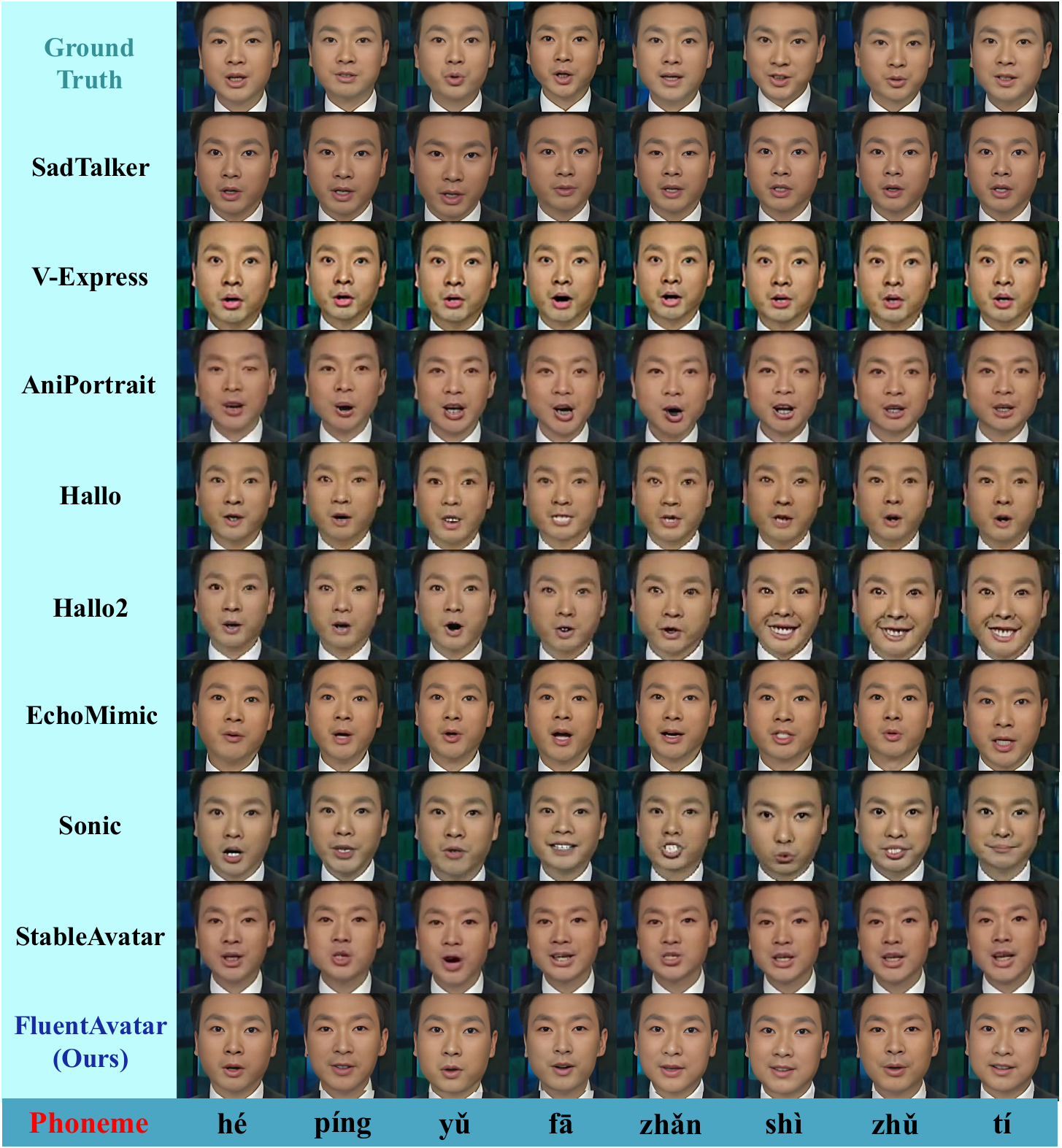}
  \caption{Qualitative comparison on CMLR and HDTF. Top: ground-truth frames. Middle: baseline results. Bottom: FluentAvatar with phoneme-to-frame alignment.}
  \label{fig:qualitative_evaluation}
  \vspace{-5pt}
\end{figure}

Qualitative comparisons in Figure~\ref{fig:qualitative_evaluation} reveal two common failure modes in existing methods. First, SadTalker, AniPortrait, and StableAvatar produce blurry reconstructions with imprecise lip articulation, while EchoMimic generates nearly static mouth shapes, indicating weak audio-visual correlation. Second, V-Express, Hallo, Hallo2, and Sonic exhibit structural degradation with warping artifacts in the lower face. In contrast, FluentAvatar generates more precise and dynamic mouth shapes that track phonemes closely while maintaining clearer facial anatomy, showing fewer artifacts and better temporal stability than existing approaches.

Beyond per-frame quality, we evaluate temporal stability via inter-frame difference heatmaps in Figure~\ref{fig:Flickering}. Diffusion-based methods exhibit widespread pixel changes across the entire frame between consecutive frames. In contrast, FluentAvatar exhibits smaller inter-frame differences that are largely confined to speech-related regions such as the mouth and jaw, more closely matching the localized motion patterns of natural speech. Moreover, this behavior is consistent with its autoregressive generation scheme, in which each frame is conditioned on previously generated content, providing a more direct constraint for modeling cross-frame dependency.

\subsection{Human Evaluation}
\begin{figure}[t]
  \centering
  \setlength{\abovecaptionskip}{1pt}
  \setlength{\belowcaptionskip}{-9pt}
  \includegraphics[width=\columnwidth]{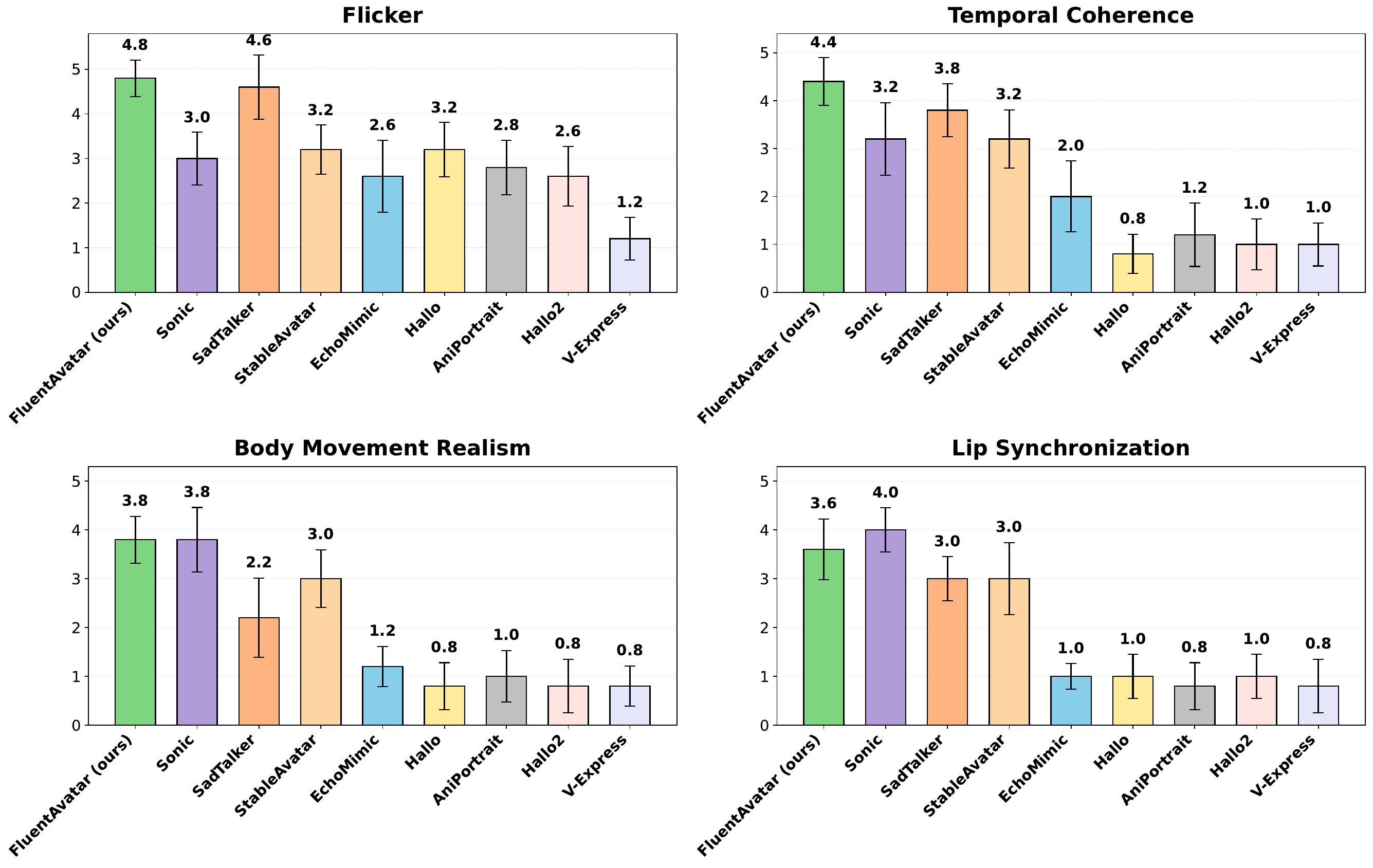}
  \caption{Human evaluation results (mean $\pm$ std).}
  \label{fig:human_eval}
\end{figure}

We conducted a human evaluation with 30 participants (15 with computer vision background, 15 general users) who rated 5 video clips per method (45 clips total) across four dimensions: flicker, body movement realism, temporal coherence, and lip synchronization, on a 5-point Likert scale. The 5 clips per method were selected to cover diverse conditions: different speakers, both Chinese and English, and varying speech tempos. Videos were randomly shuffled to prevent ordering bias. As shown in Figure~\ref{fig:human_eval}, FluentAvatar achieves the highest overall score across all four dimensions ($4.8 \pm 0.41$, $4.4 \pm 0.50$, $3.8 \pm 0.48$, $3.6 \pm 0.62$). The strongest improvements appear in flicker (4.8) and temporal coherence (4.4), directly reflecting the benefit of autoregressive sequential generation. Body movement realism (3.8) and lip synchronization (3.6) also remain competitive among the compared methods, while performance in both dimensions is not yet optimal. This limitation is related to the limited expressiveness of the discrete visual tokenizer, a known trade-off in token-based generation approaches that we discuss further in Section~\ref{sec:limitations}.

\subsection{Ablation Studies}

\paragraph{Interpolation Strategy.} To isolate the contribution of each stage, we fix the keyframes from Stage 1 and compare three interpolation configurations on CMLR (Table~\ref{tab:interp_ablation}): (A) no interpolation, directly repeating sparse keyframes; (B) replacing our Stage 2 with a diffusion-based frame interpolation model; and (C) our autoregressive timestamp-aware interpolation (full FluentAvatar). Configuration B suffers from severe flickering (BG-Flicker 0.9380 vs.\ C: 0.9985) and the worst FVD (1225.10), confirming that diffusion-based interpolation reintroduces the noise initialization problem identified in Section~\ref{sec:flickering_analysis}. Configuration A achieves trivially high BG-Flicker (repeated frames have zero difference) but poor FVD (560.28) due to the lack of continuous motion. Configuration C achieves the best FID (25.19), FVD (503.29), Sync-D (10.90), and near-GT BG-Flicker (0.9985). These results suggest that the autoregressive paradigm helps suppress flickering by avoiding per-frame noise initialization. The two-stage design is important for maintaining temporal quality and lip synchronization, while replacing either component with diffusion leads to clear performance degradation.

\paragraph{Attention Mechanisms.} We compare four attention configurations on CMLR (Table~\ref{tab:attention_ablation}): Non-Causal, Causal Accumulative, Limited History (window=2), and One-to-One (ours). Non-Causal achieves the best FID (15.63) but the worst FVD (287.95), indicating that cross-frame information leakage inflates per-frame quality while degrading temporal coherence. Our One-to-One design achieves the second-best FID (17.11) and FVD (189.24), providing the best balance between per-frame quality and temporal consistency without cross-frame leakage. Detailed definitions are in Appendix~\ref{appendix:attention_details}.

\paragraph{Additional Ablations.} We further validate three design choices on CMLR. For frame allocation (Table~\ref{tab:frame_allocation}, Appendix~\ref{app:frame_allocation}), the Dynamic strategy achieves the best FVD (503.29 vs.\ Fixed: 522.32, Random: 553.39) and Sync-D (10.90 vs.\ Fixed: 14.08), confirming that rhythm-aware allocation improves temporal coherence. For loss components (Table~\ref{tab:appendix_loss_ablation}, Appendix~\ref{appendix:loss_ablation}), the full combination (CE + LPIPS + identity + facial similarity) achieves FID 13.86 and PSNR 33.13 compared to CE-only baseline (FID 28.14, PSNR 25.08), with identity and facial similarity losses proving critical for preserving character identity. For phoneme-level conditioning (Appendix~\ref{appendix:prl}), phoneme-level units reduce face reconstruction loss by 41.8\% compared to word-level, confirming finer-grained articulatory alignment.

\begin{table}[t]
  \centering
  \caption{Interpolation strategy ablation on CMLR. All configs share identical Stage 1 keyframes; only Stage 2 differs.}
  \label{tab:interp_ablation}
  \footnotesize                   
  \setlength{\tabcolsep}{3.5pt}   
  \renewcommand{\arraystretch}{1.05}
  \resizebox{\columnwidth}{!}{%
  \begin{tabular}{lcccccc}
    \toprule
    \textbf{Config} & \textbf{FID$\downarrow$} & \textbf{FVD$\downarrow$} & \textbf{LPIPS$\downarrow$} & \textbf{PSNR$\uparrow$} & \textbf{Sync-D$\downarrow$} & \textbf{BG-Flicker$\uparrow$} \\
    \midrule
    A: Keyframes only & 33.25 & 560.28 & 0.405 & 15.02 & -- & 0.9990 \\
    B: Diffusion interp & 33.09 & 1225.10 & 0.419 & 14.87 & 13.97 & 0.9380 \\
    \textbf{C: FluentAvatar} & \textbf{25.19} & \textbf{503.29} & \textbf{0.404} & \textbf{15.07} & \textbf{10.90} & \textbf{0.9985} \\
    \bottomrule
  \end{tabular}%
  }
\end{table}

\begin{table}[t]
  \centering
  \caption{Ablation of attention mechanisms on CMLR.}
  \label{tab:attention_ablation}
  \vspace{-6pt}
  \scriptsize
  \renewcommand{\arraystretch}{1.05}
  \resizebox{\columnwidth}{!}{%
  \begin{tabular}{lcccccc}
    \toprule
    \textbf{Setting} & \textbf{FID$\downarrow$} & \textbf{FVD$\downarrow$} & \textbf{LPIPS$\downarrow$} & \textbf{PSNR$\uparrow$} & \textbf{SSIM$\uparrow$} & \textbf{Sync-C$\uparrow$} \\
    \midrule
    Non-Causal  & \textbf{15.63} & 287.95 & 0.07$\pm$0.01 & \textbf{24.41} & 0.86 & 1.11 \\
    Causal Acc. & 19.47 & 210.25 & 0.07$\pm$0.01 & 23.35 & 0.86 & \underline{1.14} \\
    Lim. Hist.  & 19.56 & \textbf{186.68} & 0.07$\pm$0.01 & 23.83 & 0.86 & \textbf{1.42} \\
    One-to-One  & \underline{17.11} & \underline{189.24} & 0.07$\pm$0.01 & \underline{24.14} & 0.86 & \textbf{1.42} \\
    \bottomrule
  \end{tabular}%
  }
\end{table}

\subsection{Limitations}
\label{sec:limitations}
Although FluentAvatar achieves strong temporal consistency, we acknowledge several limitations. First, the model relies on phonemes and timestamps extracted by ASR rather than conditioning directly on the raw audio waveform, making audio alignment indirect. Errors in phoneme extraction or timestamp estimation can propagate to video generation and lead to occasional lip-sync imprecision, which is also reflected in the human evaluation where lip synchronization (3.6) scores lower than temporal coherence (4.4). Second, due to the limited capacity of the discrete visual codebook, the visual tokenizer introduces quantization artifacts that weaken fine-grained facial details, particularly in the eye and eyebrow regions, thereby limiting subtle expression rendering. Third, our evaluation is mainly conducted on CMLR and HDTF, and generalization to more challenging scenarios, such as large head rotations, facial occlusions, and multi-person scenes, remains to be verified. In addition, BG-Flicker depends on background-region estimation obtained with DeepLabV3, so segmentation errors may introduce additional measurement bias. These limitations may be alleviated by tighter audio-phoneme-visual alignment, such as end-to-end joint optimization of ASR and generation, as well as advances in high-fidelity visual tokenizers, and more reliable evaluation.

\section{Conclusion}
Mechanism analysis and controlled experiments on diffusion-based methods indicate that denoising trajectory variation induced by stochastic initialization is an important source of inter-frame flicker. To this end, we propose FluentAvatar, a two-stage autoregressive framework built on phoneme representations, providing an effective solution to inter-frame flicker. By using phonemes as a speech-aligned intermediate representation, FluentAvatar decouples semantic alignment from visual dynamics, mitigating error accumulation and enabling fine-grained control. We further introduce BG-Flicker, a background-isolated metric for more reliable flicker evaluation in talking-head videos. Experiments on CMLR and HDTF show that FluentAvatar achieves strong performance in visual fidelity, lip synchronization, and temporal stability, with the best FVD on both benchmarks and BG-Flicker results close to ground truth. Limitations and future directions are discussed in Section~\ref{sec:limitations}.

\bibliographystyle{ACM-Reference-Format}
\bibliography{sample-base}

@String{Computer = "{IEEE} Computer" }

@String{Springer = "Springer-Verlag" }

@article{Tian2024,
  author = {Tian, K and Jiang, Y and Yuan, Z and others},
  title = {Visual autoregressive modeling: Scalable image generation via next-scale prediction},
  journal = {Advances in neural information processing systems},
  year = {2024},
  volume = {37},
  pages = {84839--84865}
}

@article{guo2024liveportrait,
  title={Liveportrait: Efficient portrait animation with stitching and retargeting control},
  author={Guo, Jianzhu and Zhang, Dingyun and Liu, Xiaoqiang and Zhong, Zhizhou and Zhang, Yuan and Wan, Pengfei and Zhang, Di},
  journal={arXiv preprint arXiv:2407.03168},
  year={2024}
}

@inproceedings{hu2024animate,
  title={Animate anyone: Consistent and controllable image-to-video synthesis for character animation},
  author={Hu, Li},
  booktitle={Proceedings of the IEEE/CVF Conference on Computer Vision and Pattern Recognition},
  pages={8153--8163},
  year={2024}
}

@inproceedings{tian2024emo,
  title={Emo: Emote portrait alive generating expressive portrait videos with audio2video diffusion model under weak conditions},
  author={Tian, Linrui and Wang, Qi and Zhang, Bang and Bo, Liefeng},
  booktitle={European Conference on Computer Vision},
  pages={244--260},
  year={2024},
  organization={Springer}
}

@inproceedings{chen2025echomimic,
  title={Echomimic: Lifelike audio-driven portrait animations through editable landmark conditions},
  author={Chen, Zhiyuan and Cao, Jiajiong and Chen, Zhiquan and Li, Yuming and Ma, Chenguang},
  booktitle={Proceedings of the AAAI Conference on Artificial Intelligence},
  volume={39},
  pages={2403--2410},
  year={2025}
}

@inproceedings{wang2021one,
  title={One-shot free-view neural talking-head synthesis for video conferencing},
  author={Wang, Ting-Chun and Mallya, Arun and Liu, Ming-Yu},
  booktitle={Proceedings of the IEEE/CVF conference on computer vision and pattern recognition},
  pages={10039--10049},
  year={2021}
}

@article{ho2020denoising,
  title={Denoising diffusion probabilistic models},
  author={Ho, Jonathan and Jain, Ajay and Abbeel, Pieter},
  journal={Advances in neural information processing systems},
  volume={33},
  pages={6840--6851},
  year={2020}
}

@inproceedings{prajwal2020lip,
  title={A lip sync expert is all you need for speech to lip generation in the wild},
  author={Prajwal, KR and Mukhopadhyay, Rudrabha and Namboodiri, Vinay P and Jawahar, CV},
  booktitle={Proceedings of the 28th ACM international conference on multimedia},
  pages={484--492},
  year={2020}
}

@inproceedings{zhang2023sadtalker,
  title={Sadtalker: Learning realistic 3d motion coefficients for stylized audio-driven single image talking face animation},
  author={Zhang, Wenxuan and Cun, Xiaodong and Wang, Xuan and Zhang, Yong and Shen, Xi and Guo, Yu and Shan, Ying and Wang, Fei},
  booktitle={Proceedings of the IEEE/CVF conference on computer vision and pattern recognition},
  pages={8652--8661},
  year={2023}
}

@article{xu2024hallo,
  title={Hallo: Hierarchical audio-driven visual synthesis for portrait image animation},
  author={Xu, Mingwang and Li, Hui and Su, Qingkun and Shang, Hanlin and Zhang, Liwei and Liu, Ce and Wang, Jingdong and Yao, Yao and Zhu, Siyu},
  journal={arXiv preprint arXiv:2406.08801},
  year={2024}
}

@article{cui2024hallo2,
  title={Hallo2: Long-duration and high-resolution audio-driven portrait image animation},
  author={Cui, Jiahao and Li, Hui and Yao, Yao and Zhu, Hao and Shang, Hanlin and Cheng, Kaihui and Zhou, Hang and Zhu, Siyu and Wang, Jingdong},
  journal={arXiv preprint arXiv:2410.07718},
  year={2024}
}

@article{wang2024v,
  title={V-express: Conditional dropout for progressive training of portrait video generation},
  author={Wang, Cong and Tian, Kuan and Zhang, Jun and Guan, Yonghang and Luo, Feng and Shen, Fei and Jiang, Zhiwei and Gu, Qing and Han, Xiao and Yang, Wei},
  journal={arXiv preprint arXiv:2406.02511},
  year={2024}
}

@article{ji2024sonic,
  title={Sonic: Shifting Focus to Global Audio Perception in Portrait Animation},
  author={Ji, Xiaozhong and Hu, Xiaobin and Xu, Zhihong and Zhu, Junwei and Lin, Chuming and He, Qingdong and Zhang, Jiangning and Luo, Donghao and Chen, Yi and Lin, Qin and others},
  journal={arXiv preprint arXiv:2411.16331},
  year={2024}
}

@article{lin2025omnihuman,
  title={OmniHuman-1: Rethinking the Scaling-Up of One-Stage Conditioned Human Animation Models},
  author={Lin, Gaojie and Jiang, Jianwen and Yang, Jiaqi and Zheng, Zerong and Liang, Chao},
  journal={arXiv preprint arXiv:2502.01061},
  year={2025}
}

@article{li2024latentsync,
  title={LatentSync: Audio Conditioned Latent Diffusion Models for Lip Sync},
  author={Li, Chunyu and Zhang, Chao and Xu, Weikai and Xie, Jinghui and Feng, Weiguo and Peng, Bingyue and Xing, Weiwei},
  journal={arXiv preprint arXiv:2412.09262},
  year={2024}
}

@article{jiang2024loopy,
  title={Loopy: Taming audio-driven portrait avatar with long-term motion dependency},
  author={Jiang, Jianwen and Liang, Chao and Yang, Jiaqi and Lin, Gaojie and Zhong, Tianyun and Zheng, Yanbo},
  journal={arXiv preprint arXiv:2409.02634},
  year={2024}
}

@inproceedings{chen2019hierarchical,
  title={Hierarchical cross-modal talking face generation with dynamic pixel-wise loss},
  author={Chen, Lele and Maddox, Ross K and Duan, Zhiyao and Xu, Chenliang},
  booktitle={Proceedings of the IEEE/CVF conference on computer vision and pattern recognition},
  pages={7832--7841},
  year={2019}
}

@inproceedings{das2020speech,
  title={Speech-driven facial animation using cascaded gans for learning of motion and texture},
  author={Das, Dipanjan and Biswas, Sandika and Sinha, Sanjana and Bhowmick, Brojeshwar},
  booktitle={Computer Vision--ECCV 2020: 16th European Conference, Glasgow, UK, August 23--28, 2020, Proceedings, Part XXX 16},
  pages={408--424},
  year={2020},
  organization={Springer}
}

@inproceedings{meshry2021learned,
  title={Learned spatial representations for few-shot talking-head synthesis},
  author={Meshry, Moustafa and Suri, Saksham and Davis, Larry S and Shrivastava, Abhinav},
  booktitle={Proceedings of the IEEE/CVF international conference on computer vision},
  pages={13829--13838},
  year={2021}
}

@inproceedings{zhou2019talking,
  title={Talking face generation by adversarially disentangled audio-visual representation},
  author={Zhou, Hang and Liu, Yu and Liu, Ziwei and Luo, Ping and Wang, Xiaogang},
  booktitle={Proceedings of the AAAI conference on artificial intelligence},
  volume={33},
  pages={9299--9306},
  year={2019}
}

@inproceedings{zhou2021pose,
  title={Pose-controllable talking face generation by implicitly modularized audio-visual representation},
  author={Zhou, Hang and Sun, Yasheng and Wu, Wayne and Loy, Chen Change and Wang, Xiaogang and Liu, Ziwei},
  booktitle={Proceedings of the IEEE/CVF conference on computer vision and pattern recognition},
  pages={4176--4186},
  year={2021}
}

@article{xu2024vasa,
  title={Vasa-1: Lifelike audio-driven talking faces generated in real time},
  author={Xu, Sicheng and Chen, Guojun and Guo, Yu-Xiao and Yang, Jiaolong and Li, Chong and Zang, Zhenyu and Zhang, Yizhong and Tong, Xin and Guo, Baining},
  journal={Advances in Neural Information Processing Systems},
  volume={37},
  pages={660--684},
  year={2024}
}

@article{zhang2024musetalk,
  title={Musetalk: Real-time high quality lip synchronization with latent space inpainting},
  author={Zhang, Yue and Liu, Minhao and Chen, Zhaokang and Wu, Bin and Zeng, Yubin and Zhan, Chao and He, Yingjie and Huang, Junxin and Zhou, Wenjiang},
  journal={arXiv preprint arXiv:2410.10122},
  year={2024}
}

@article{yan2021videogpt,
  title={Videogpt: Video generation using vq-vae and transformers},
  author={Yan, Wilson and Zhang, Yunzhi and Abbeel, Pieter and Srinivas, Aravind},
  journal={arXiv preprint arXiv:2104.10157},
  year={2021}
}

@inproceedings{yu2023magvit,
  title={Magvit: Masked generative video transformer},
  author={Yu, Lijun and Cheng, Yong and Sohn, Kihyuk and Lezama, Jos{\'e} and Zhang, Han and Chang, Huiwen and Hauptmann, Alexander G and Yang, Ming-Hsuan and Hao, Yuan and Essa, Irfan and others},
  booktitle={Proceedings of the IEEE/CVF Conference on Computer Vision and Pattern Recognition},
  pages={10459--10469},
  year={2023}
}

@article{achiam2023gpt,
  title={Gpt-4 technical report},
  author={Achiam, Josh and Adler, Steven and Agarwal, Sandhini and Ahmad, Lama and Akkaya, Ilge and Aleman, Florencia Leoni and Almeida, Diogo and Altenschmidt, Janko and Altman, Sam and Anadkat, Shyamal and others},
  journal={arXiv preprint arXiv:2303.08774},
  year={2023}
}

@article{touvron2023llama,
  title={Llama: Open and efficient foundation language models},
  author={Touvron, Hugo and Lavril, Thibaut and Izacard, Gautier and Martinet, Xavier and Lachaux, Marie-Anne and Lacroix, Timoth{\'e}e and Rozi{\`e}re, Baptiste and Goyal, Naman and Hambro, Eric and Azhar, Faisal and others},
  journal={arXiv preprint arXiv:2302.13971},
  year={2023}
}

@inproceedings{liang2024survey,
  title={A Survey of Multimodel Large Language Models},
  author={Liang, Zijing and Xu, Yanjie and Hong, Yifan and Shang, Penghui and Wang, Qi and Fu, Qiang and Liu, Ke},
  booktitle={Proceedings of the 3rd International Conference on Computer, Artificial Intelligence and Control Engineering},
  pages={405--409},
  year={2024}
}

@inproceedings{chang2022maskgit,
  title={Maskgit: Masked generative image transformer},
  author={Chang, Huiwen and Zhang, Han and Jiang, Lu and Liu, Ce and Freeman, William T},
  booktitle={Proceedings of the IEEE/CVF conference on computer vision and pattern recognition},
  pages={11315--11325},
  year={2022}
}

@article{sun2024autoregressive,
  title={Autoregressive model beats diffusion: Llama for scalable image generation},
  author={Sun, Peize and Jiang, Yi and Chen, Shoufa and Zhang, Shilong and Peng, Bingyue and Luo, Ping and Yuan, Zehuan},
  journal={arXiv preprint arXiv:2406.06525},
  year={2024}
}

@article{kondratyuk2023videopoet,
  title={Videopoet: A large language model for zero-shot video generation},
  author={Kondratyuk, Dan and Yu, Lijun and Gu, Xiuye and Lezama, Jos{\'e} and Huang, Jonathan and Schindler, Grant and Hornung, Rachel and Birodkar, Vighnesh and Yan, Jimmy and Chiu, Ming-Chang and others},
  journal={arXiv preprint arXiv:2312.14125},
  year={2023}
}

@article{xie2024show,
  title={Show-o: One single transformer to unify multimodal understanding and generation},
  author={Xie, Jinheng and Mao, Weijia and Bai, Zechen and Zhang, David Junhao and Wang, Weihao and Lin, Kevin Qinghong and Gu, Yuchao and Chen, Zhijie and Yang, Zhenheng and Shou, Mike Zheng},
  journal={arXiv preprint arXiv:2408.12528},
  year={2024}
}

@article{wang2024emu3,
  title={Emu3: Next-token prediction is all you need},
  author={Wang, Xinlong and Zhang, Xiaosong and Luo, Zhengxiong and Sun, Quan and Cui, Yufeng and Wang, Jinsheng and Zhang, Fan and Wang, Yueze and Li, Zhen and Yu, Qiying and others},
  journal={arXiv preprint arXiv:2409.18869},
  year={2024}
}

@article{wei2024aniportrait,
  title={Aniportrait: Audio-driven synthesis of photorealistic portrait animation},
  author={Wei, Huawei and Yang, Zejun and Wang, Zhisheng},
  journal={arXiv preprint arXiv:2403.17694},
  year={2024}
}

@article{hong2022cogvideo,
  title={Cogvideo: Large-scale pretraining for text-to-video generation via transformers},
  author={Hong, Wenyi and Ding, Ming and Zheng, Wendi and Liu, Xinghan and Tang, Jie},
  journal={arXiv preprint arXiv:2205.15868},
  year={2022}
}

@article{luo2024open,
  title={Open-magvit2: An open-source project toward democratizing auto-regressive visual generation},
  author={Luo, Zhuoyan and Shi, Fengyuan and Ge, Yixiao and Yang, Yujiu and Wang, Limin and Shan, Ying},
  journal={arXiv preprint arXiv:2409.04410},
  year={2024}
}

@inproceedings{zhao2019cascade,
  title={A cascade sequence-to-sequence model for chinese mandarin lip reading},
  author={Zhao, Ya and Xu, Rui and Song, Mingli},
  booktitle={Proceedings of the 1st ACM International Conference on Multimedia in Asia},
  pages={1--6},
  year={2019}
}

@article{chu2025artalk,
  title={ARTalk: Speech-Driven 3D Head Animation via Autoregressive Model},
  author={Chu, Xuangeng and Goswami, Nabarun and Cui, Ziteng and Wang, Hanqin and Harada, Tatsuya},
  journal={arXiv preprint arXiv:2502.20323},
  year={2025}
}

@inproceedings{zhen2025teller,
  title={Teller: Real-Time Streaming Audio-Driven Portrait Animation with Autoregressive Motion Generation},
  author={Zhen, Dingcheng and Yin, Shunshun and Qin, Shiyang and Yi, Hou and Zhang, Ziwei and Liu, Siyuan and Qi, Gan and Tao, Ming},
  booktitle={Proceedings of the Computer Vision and Pattern Recognition Conference},
  pages={21075--21085},
  year={2025}
}

@inproceedings{wang2025instructavatar,
  title={Instructavatar: Text-guided emotion and motion control for avatar generation},
  author={Wang, Yuchi and Guo, Junliang and Bai, Jianhong and Yu, Runyi and He, Tianyu and Tan, Xu and Sun, Xu and Bian, Jiang},
  booktitle={Proceedings of the AAAI Conference on Artificial Intelligence},
  volume={39},
  pages={8132--8140},
  year={2025}
}

@article{zhang2024vfimamba,
  title={Vfimamba: Video frame interpolation with state space models},
  author={Zhang, Guozhen and Liu, Chuxnu and Cui, Yutao and Zhao, Xiaotong and Ma, Kai and Wang, Limin},
  journal={Advances in Neural Information Processing Systems},
  volume={37},
  pages={107225--107248},
  year={2024}
}

@inproceedings{zhang2021flow,
  title={Flow-guided one-shot talking face generation with a high-resolution audio-visual dataset},
  author={Zhang, Zhimeng and Li, Lincheng and Ding, Yu and Fan, Changjie},
  booktitle={Proceedings of the IEEE/CVF conference on computer vision and pattern recognition},
  pages={3661--3670},
  year={2021}
}

@inproceedings{zhang2018unreasonable,
  title={The unreasonable effectiveness of deep features as a perceptual metric},
  author={Zhang, Richard and Isola, Phillip and Efros, Alexei A and Shechtman, Eli and Wang, Oliver},
  booktitle={Proceedings of the IEEE conference on computer vision and pattern recognition},
  pages={586--595},
  year={2018}
}

@inproceedings{peebles2023scalable,
  title={Scalable diffusion models with transformers},
  author={Peebles, William and Xie, Saining},
  booktitle={Proceedings of the IEEE/CVF international conference on computer vision},
  pages={4195--4205},
  year={2023}
}

@article{tu2025stableavatar,
  title={Stableavatar: Infinite-length audio-driven avatar video generation},
  author={Tu, Shuyuan and Pan, Yueming and Huang, Yinming and Han, Xintong and Xing, Zhen and Dai, Qi and Luo, Chong and Wu, Zuxuan and Jiang, Yu-Gang},
  journal={arXiv preprint arXiv:2508.08248},
  year={2025}
}

@article{harvey2022flexible,
  title={Flexible diffusion modeling of long videos},
  author={Harvey, William and Naderiparizi, Saeid and Masrani, Vaden and Weilbach, Christian and Wood, Frank},
  journal={Advances in neural information processing systems},
  volume={35},
  pages={27953--27965},
  year={2022}
}

@article{zhang2024impact,
  title={Impact of using virtual avatars in educational videos on user experience},
  author={Zhang, Ruyuan and Wu, Qun},
  journal={Scientific Reports},
  volume={14},
  number={1},
  pages={6592},
  year={2024}
}

@inproceedings{huang2024vbench,
  title={Vbench: Comprehensive benchmark suite for video generative models},
  author={Huang, Ziqi and He, Yinan and Yu, Jiashuo and Zhang, Fan and Si, Chenyang and Jiang, Yuming and Zhang, Yuanhan and Wu, T. and Jin, Q. and Chanpaisit, N. and Wang, Y. and Liu, Z.},
  booktitle={Proceedings of the IEEE/CVF Conference on Computer Vision and Pattern Recognition},
  pages={21807--21818},
  year={2024}
}

@article{chen2017rethinking,
  title={Rethinking atrous convolution for semantic image segmentation},
  author={Chen, Liang-Chieh and Papandreou, George and Schroff, Florian and Adam, Hartwig},
  journal={arXiv preprint arXiv:1706.05587},
  year={2017}
}

@inproceedings{ge2023preserve,
  title={Preserve your own correlation: A noise prior for video diffusion models},
  author={Ge, Songwei and Nah, Seungjun and Liu, Guilin and Poon, Tyler and Tao, Andrew and Catanzaro, Bryan and Jacobs, David and Huang, Jia-Bin and Liu, Ming-Yu and Balaji, Yogesh},
  booktitle={Proceedings of the IEEE/CVF International Conference on Computer Vision},
  pages={22930--22941},
  year={2023}
}

@article{huang2025self,
  title={Self forcing: Bridging the train-test gap in autoregressive video diffusion},
  author={Huang, Xun and Li, Zhengqi and He, Guande and Zhou, Mingyuan and Shechtman, Eli},
  journal={arXiv preprint arXiv:2506.08009},
  year={2025}
}

@inproceedings{guo2025arig,
  title={Arig: Autoregressive interactive head generation for real-time conversations},
  author={Guo, Ying and Liu, Xi and Zhen, Cheng and Yan, Pengfei and Wei, Xiaoming},
  booktitle={Proceedings of the IEEE/CVF International Conference on Computer Vision},
  pages={12956--12965},
  year={2025}
}

@article{huang2025live,
  title={Live avatar: Streaming real-time audio-driven avatar generation with infinite length},
  author={Huang, Yubo and Guo, Hailong and Wu, Fangtai and Zhang, Shifeng and Huang, Shijie and Gan, Qijun and Liu, Lin and Zhao, Sirui and Chen, Enhong and Liu, Jiaming and others},
  journal={arXiv preprint arXiv:2512.04677},
  year={2025}
}

@inproceedings{li2025ditto,
  title={Ditto: Motion-space diffusion for controllable realtime talking head synthesis},
  author={Li, Tianqi and Zheng, Ruobing and Yang, Minghui and Chen, Jingdong and Yang, Ming},
  booktitle={Proceedings of the 33rd ACM International Conference on Multimedia},
  pages={9704--9713},
  year={2025}
}

@article{gan2025omniavatar,
  title={Omniavatar: Efficient audio-driven avatar video generation with adaptive body animation},
  author={Gan, Qijun and Yang, Ruizi and Zhu, Jianke and Xue, Shaofei and Hoi, Steven},
  journal={arXiv preprint arXiv:2506.18866},
  year={2025}
}

@article{tian2024visual,
  title={Visual autoregressive modeling: Scalable image generation via next-scale prediction},
  author={Tian, Keyu and Jiang, Yi and Yuan, Zehuan and Peng, Bingyue and Wang, Liwei},
  journal={Advances in neural information processing systems},
  volume={37},
  pages={84839--84865},
  year={2024}
}

\clearpage
\appendix

\section{Data Preparation Details}
\label{appendix:data_preparation}

\subsection{CMLR Super-Resolution}
\label{appendix:cmlr_super-resolution}

\begin{figure}[h]
  \centering
  \includegraphics[width=\columnwidth]{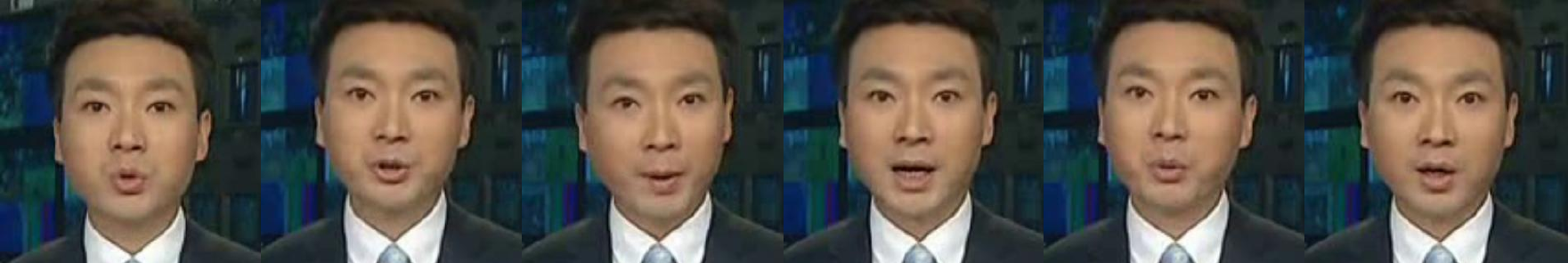}
  \vspace{2pt}
  \centerline{\small (a) Original video frames}
  \vspace{6pt}
  \includegraphics[width=\columnwidth]{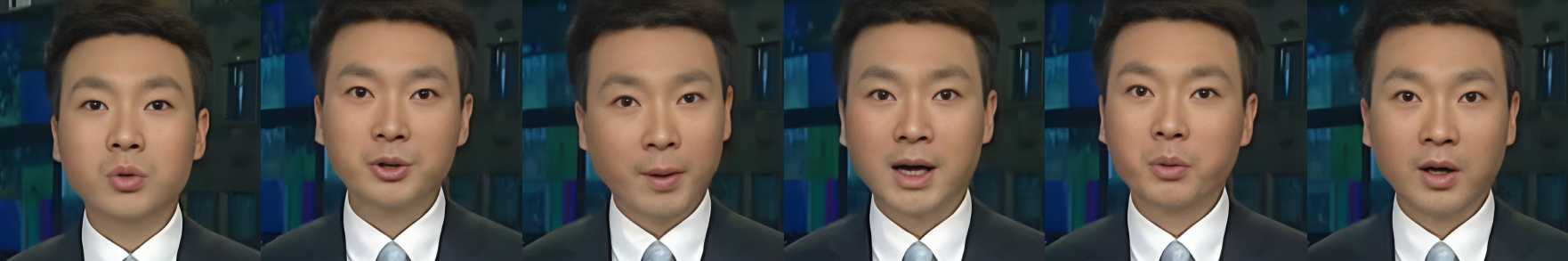}
  \vspace{2pt}
  \centerline{\small (b) Enhanced frames after $4\times$ super-resolution}
  \caption{CMLR dataset before and after super-resolution.}
  \label{fig:cmlr_sr}
\end{figure}

The scarcity of high-quality, large-scale Chinese talking-head datasets poses a significant challenge to research in this domain. The CMLR dataset is one of the few publicly available Chinese datasets for this task, but its low resolution results in blurry facial features and insufficient lip detail, compromising both training efficacy and evaluation reliability. To address this, we apply GFPGAN for $4\times$ super-resolution across the entire CMLR dataset. Figure~\ref{fig:cmlr_sr} shows a visual comparison before and after enhancement. We will open-source this enhanced dataset to benefit the community.

\subsection{Face Cropping Strategy}
\label{appendix:crop_visuals}

\begin{figure}[h]
  \centering
  \includegraphics[width=\columnwidth]{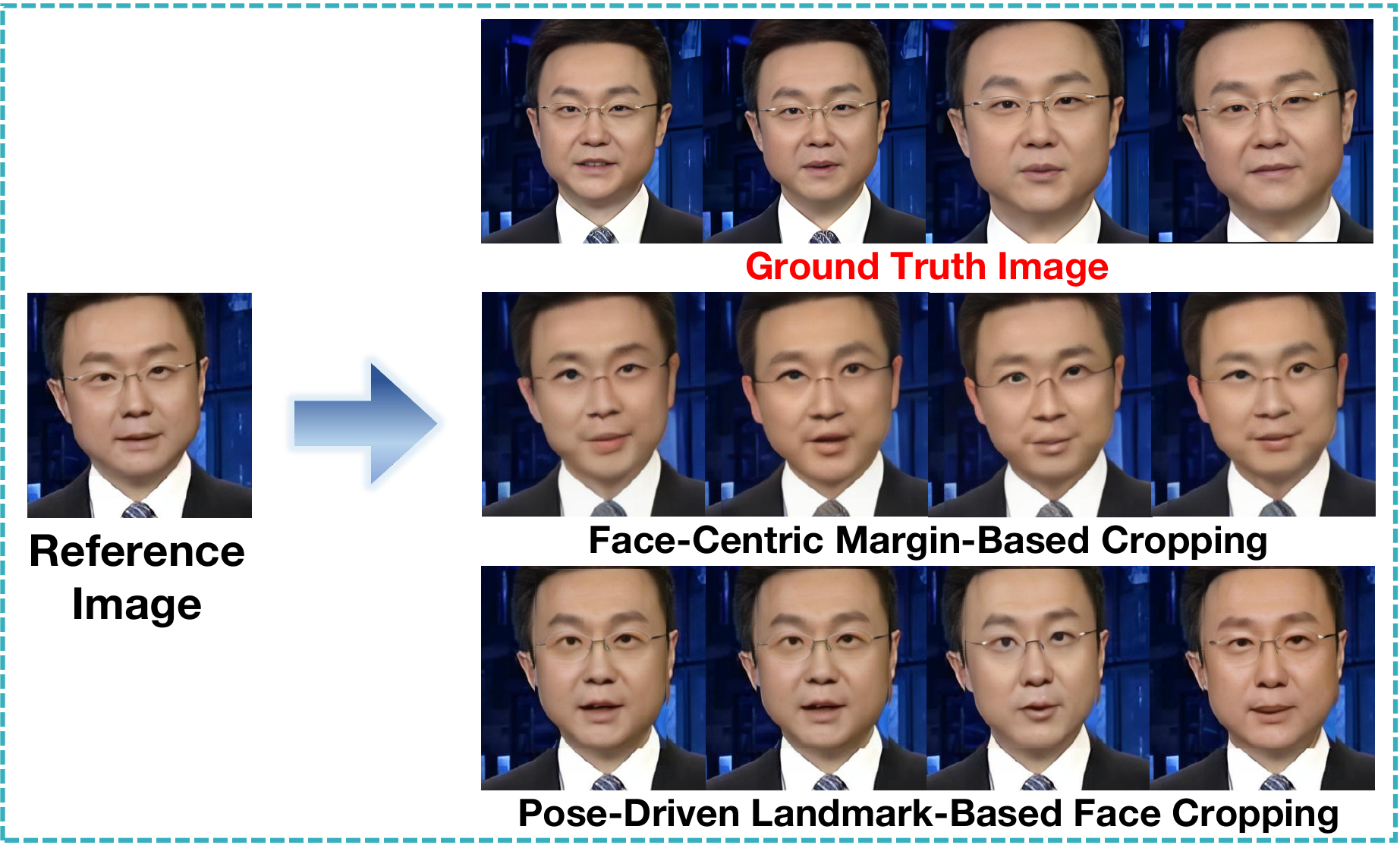}
  \caption{Face-Centric Cropping results.}
  \label{fig:face_crop_1}
\end{figure}

\begin{figure}[t]
  \centering
  \includegraphics[width=\columnwidth]{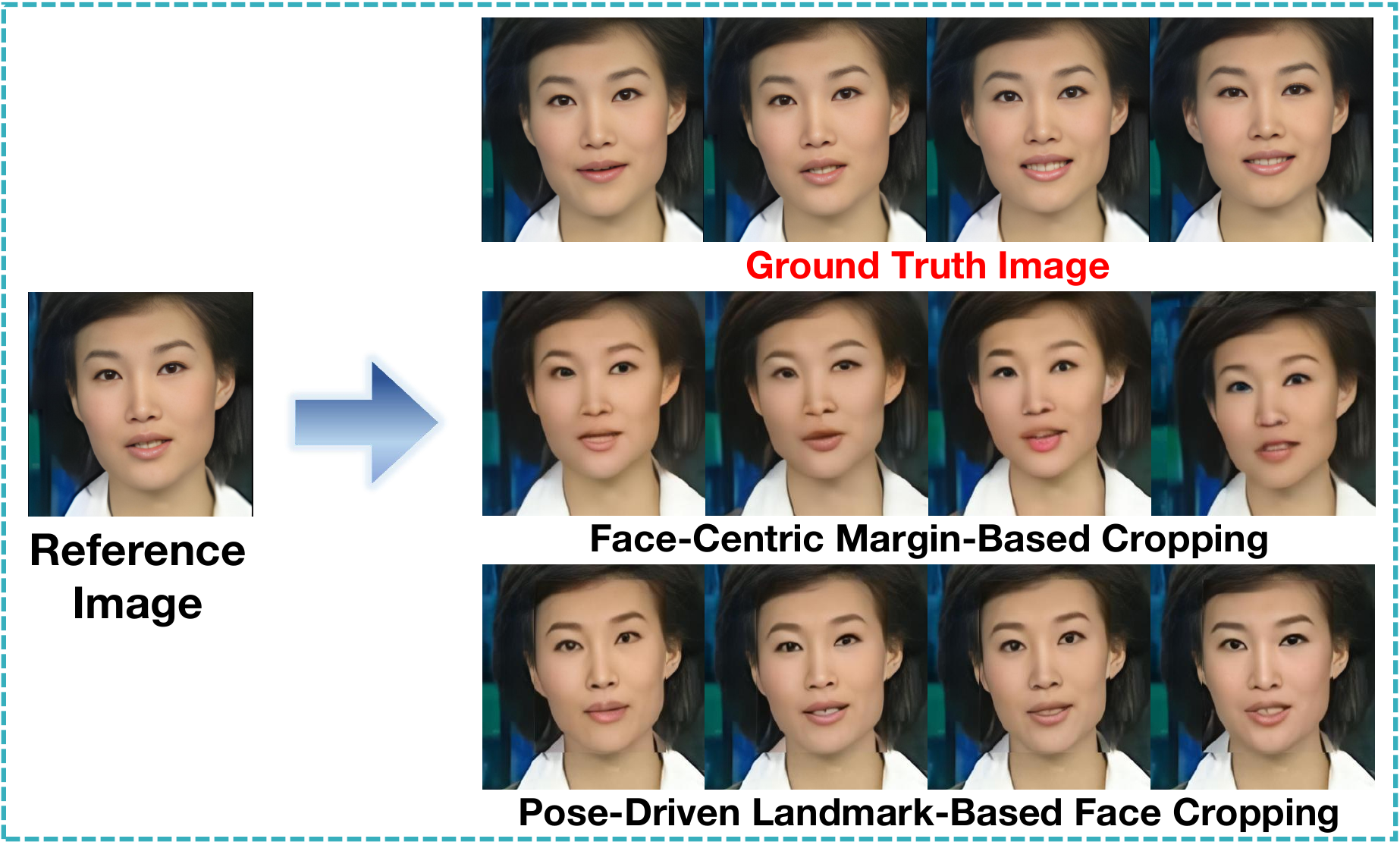}
  \caption{Pose-Driven Landmark Cropping results.}
  \label{fig:face_crop_2}
\end{figure}

We compare two face preprocessing methods: Face-Centric Cropping and Pose-Driven Landmark Cropping. The former leads to unstable generation due to scale and background variations, while the landmark-based approach ensures tighter alignment and more consistent lip dynamics. As shown in Figure~\ref{fig:face_crop_1} and Figure~\ref{fig:face_crop_2}, our Pose-Driven method yields tighter facial alignment and more consistent lip dynamics. Quantitatively, Table~\ref{tab:appendix_avg_similarity_comparison} confirms it achieves lower (better) Identity Similarity Scores on 3 out of 4 face recognition models. We adopt Pose-Driven Landmark Cropping for all experiments.

\begin{table*}[t]
\centering
\caption{Identity Similarity Score (ISS) under different cropping strategies. Lower is better. \textbf{Bold} in the Total row indicates the better strategy per model.}
\label{tab:appendix_avg_similarity_comparison}
\small
\setlength{\tabcolsep}{5pt}
\renewcommand{\arraystretch}{1.1}
\begin{tabular}{lcccc|cccc}
  \toprule
  \multirow{2}{*}{\textbf{Subset}}
  & \multicolumn{4}{c|}{\textbf{Face-Centric Cropping}}
  & \multicolumn{4}{c}{\textbf{Landmark-Based Cropping}} \\
  & ArcFace & FaceNet & FaceNet512 & VGG-Face
  & ArcFace & FaceNet & FaceNet512 & VGG-Face \\
  \midrule
  s1 & 0.296 & 0.161 & 0.193 & 0.290 & 0.325 & 0.218 & 0.236 & 0.340 \\
  s2 & 0.219 & 0.167 & 0.128 & 0.289 & 0.208 & 0.189 & 0.101 & 0.224 \\
  s3 & 0.258 & 0.172 & 0.108 & 0.290 & 0.287 & 0.178 & 0.075 & 0.201 \\
  s4 & 0.370 & 0.342 & 0.220 & 0.364 & 0.363 & 0.282 & 0.192 & 0.347 \\
  s5 & 0.314 & 0.279 & 0.168 & 0.359 & 0.302 & 0.198 & 0.132 & 0.263 \\
  \midrule
  \textbf{Total} & \textbf{0.291} & 0.224 & 0.163 & 0.318 & 0.297 & \textbf{0.213} & \textbf{0.147} & \textbf{0.275} \\
  \bottomrule
\end{tabular}
\end{table*}

\section{Training Details}
\label{appendix:training_process}

\begin{figure}[t]
  \centering
  \begin{minipage}[t]{0.48\columnwidth}
    \centering
    \includegraphics[width=\linewidth]{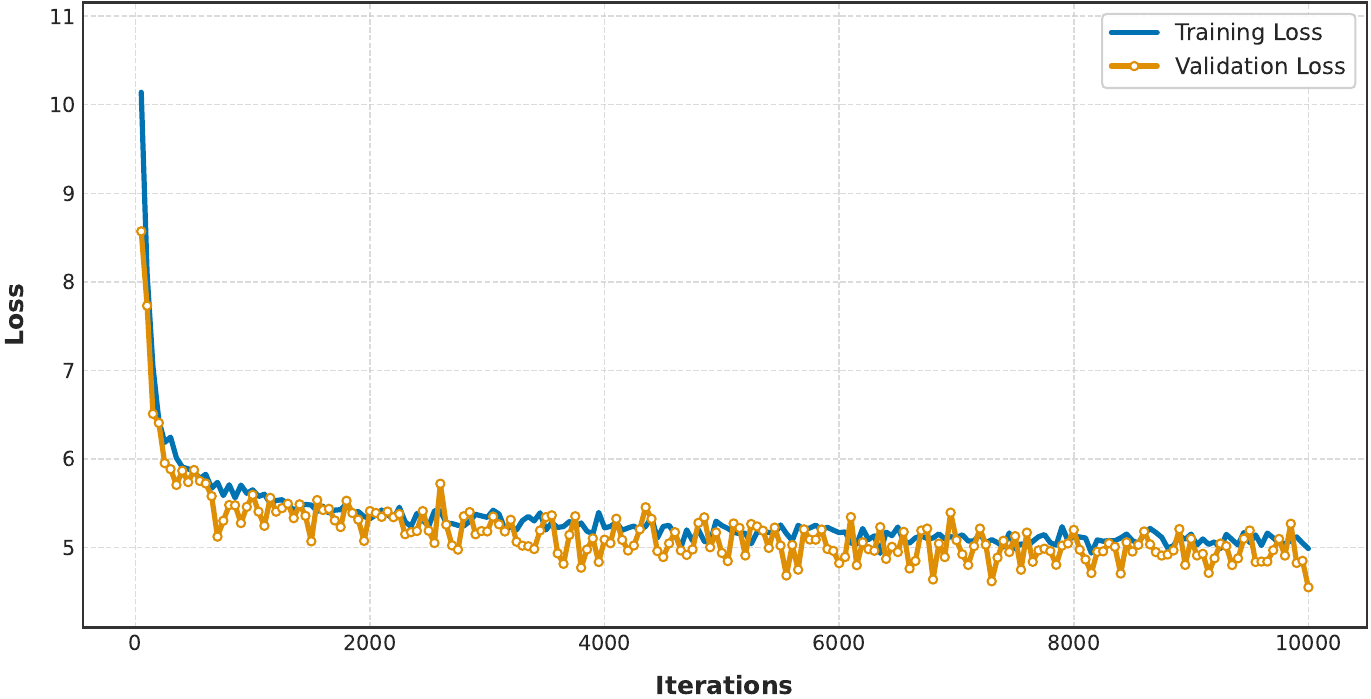}
    \centerline{\small (a) Total loss}
  \end{minipage}
  \hfill
  \begin{minipage}[t]{0.48\columnwidth}
    \centering
    \includegraphics[width=\linewidth]{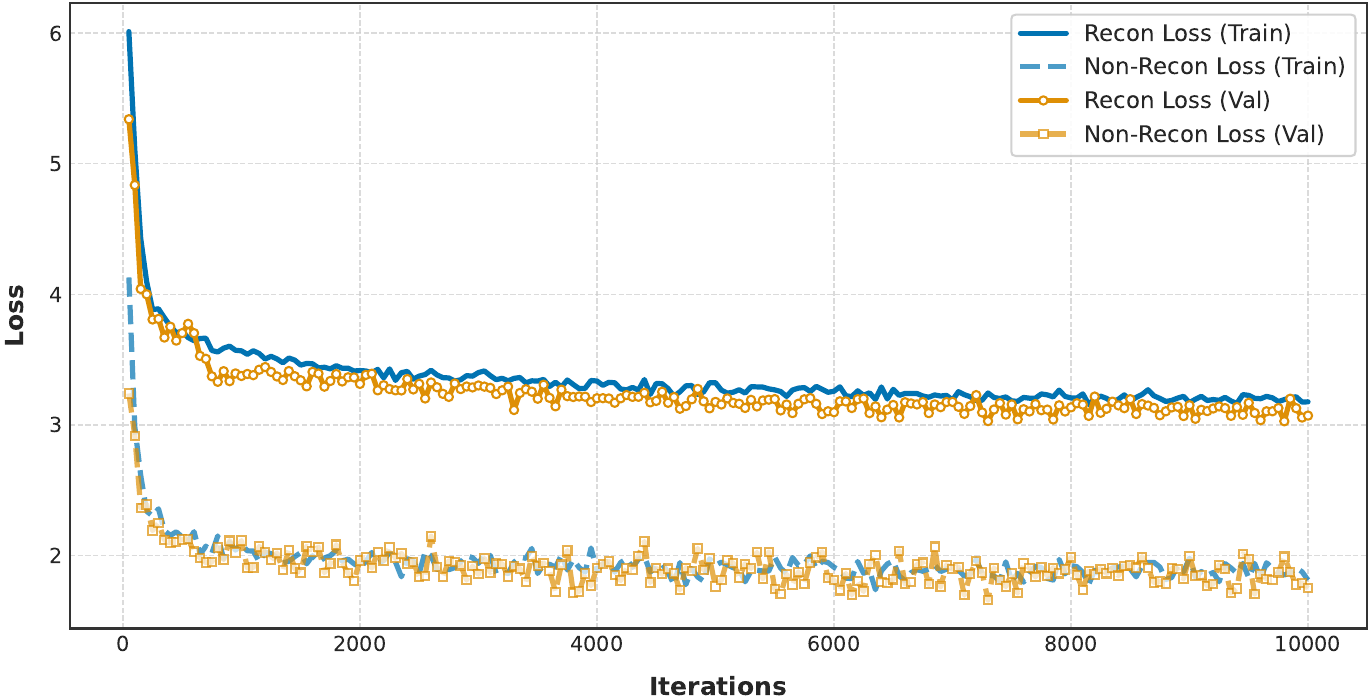}
    \centerline{\small (b) Recon vs.\ non-recon}
  \end{minipage}
  \caption{Training loss curves (1/3): total loss and reconstruction vs.\ non-reconstruction loss over 10,000 steps.}
  \label{fig:loss_row1}
\end{figure}

\begin{figure}[t]
  \centering
  \begin{minipage}[t]{0.48\columnwidth}
    \centering
    \includegraphics[width=\linewidth]{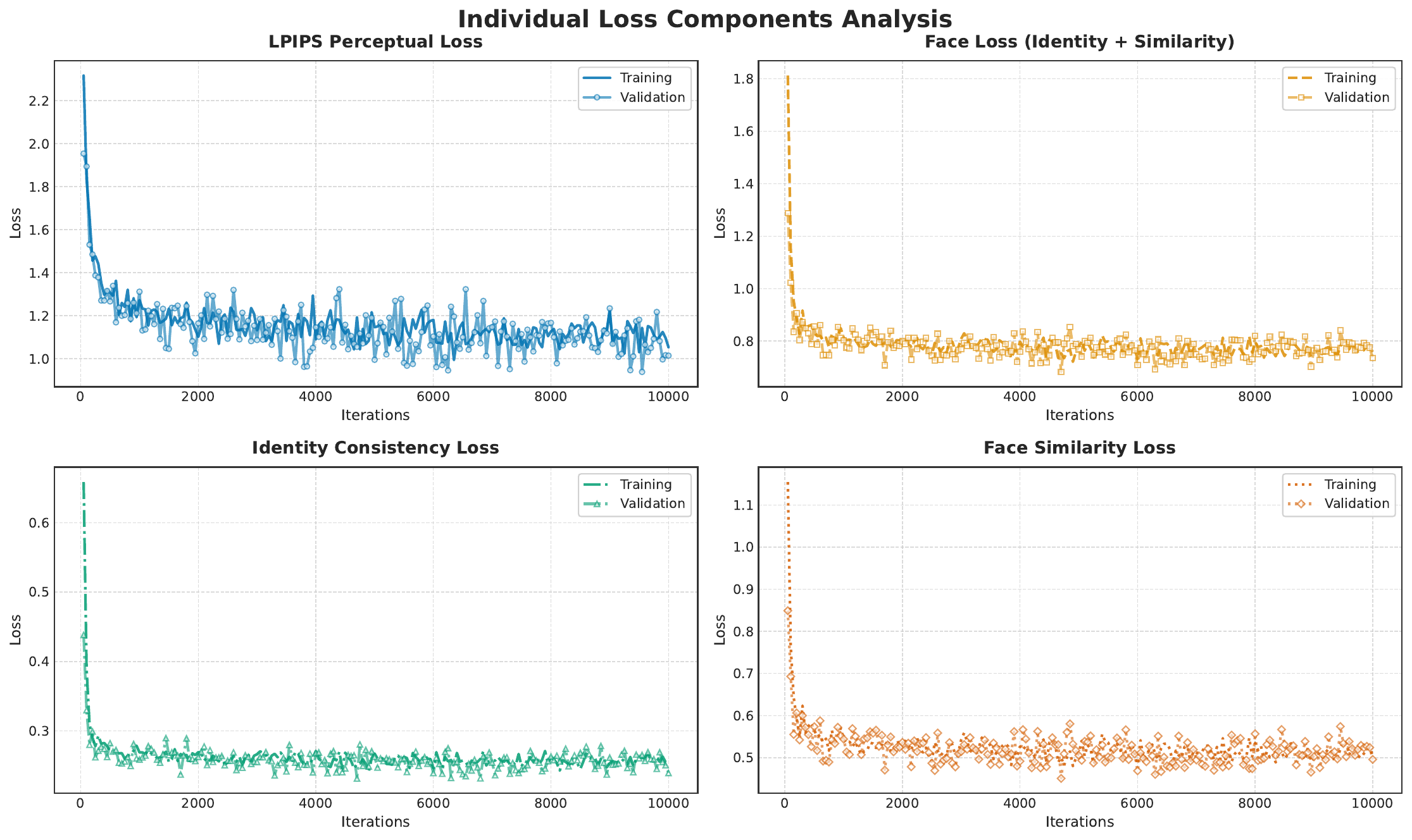}
    \centerline{\small (a) Individual losses}
  \end{minipage}
  \hfill
  \begin{minipage}[t]{0.48\columnwidth}
    \centering
    \includegraphics[width=\linewidth]{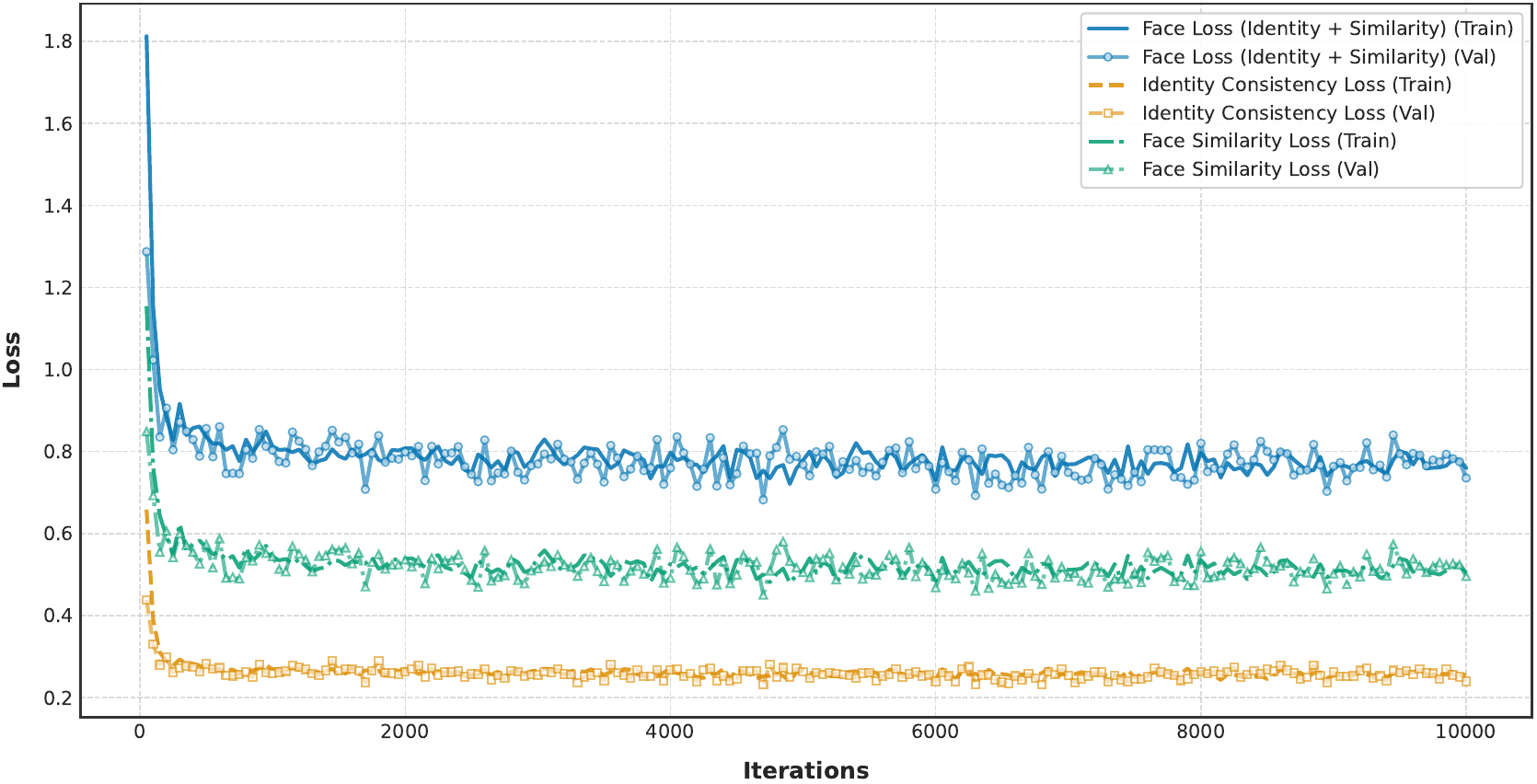}
    \centerline{\small (b) Face-related losses}
  \end{minipage}
  \caption{Training loss curves (2/3): individual loss components and face-specific losses.}
  \label{fig:loss_row2}
\end{figure}

\begin{figure}[t]
  \centering
  \begin{minipage}[t]{0.48\columnwidth}
    \centering
    \includegraphics[width=\linewidth]{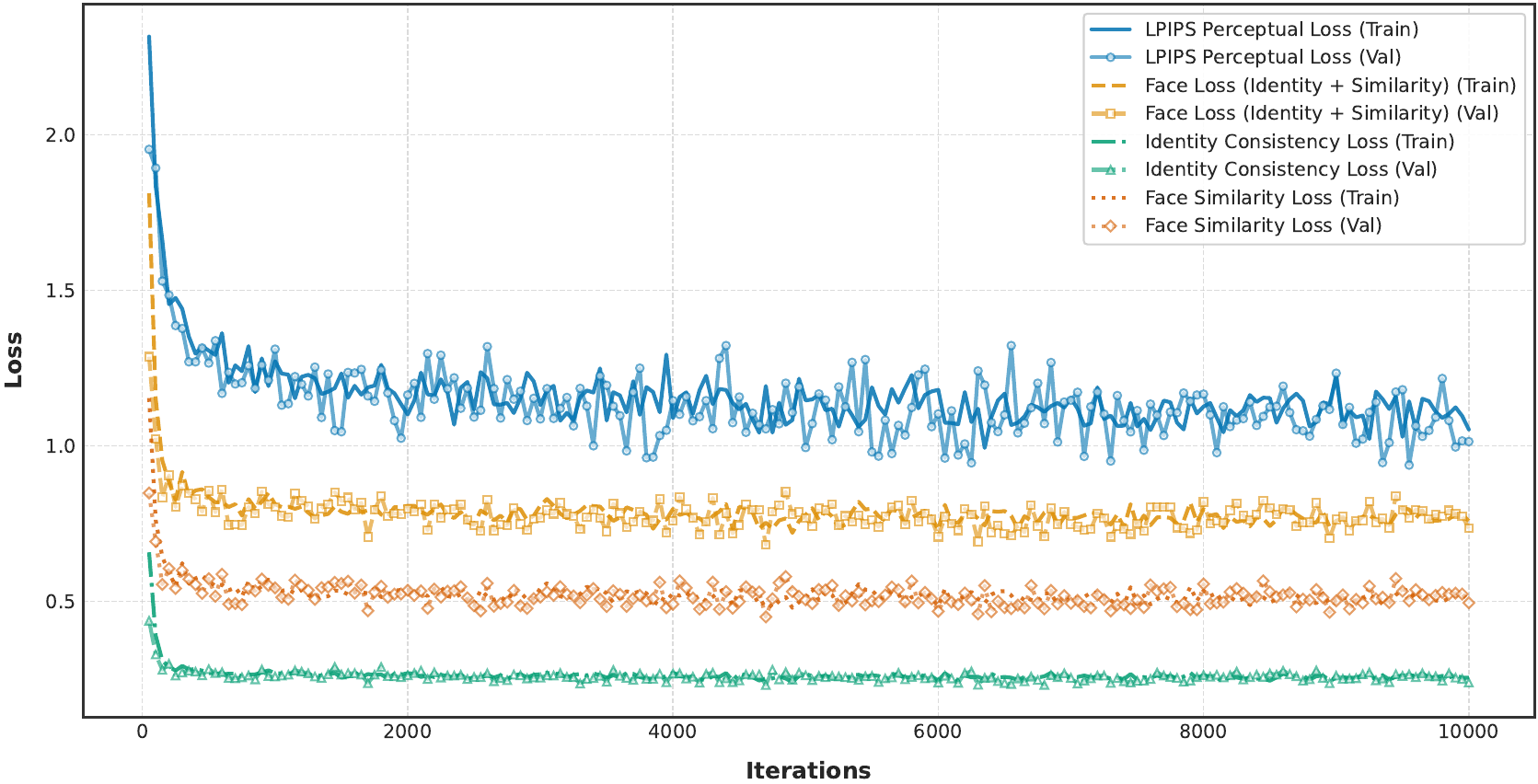}
    \centerline{\small (a) Detailed losses}
  \end{minipage}
  \hfill
  \begin{minipage}[t]{0.48\columnwidth}
    \centering
    \includegraphics[width=\linewidth]{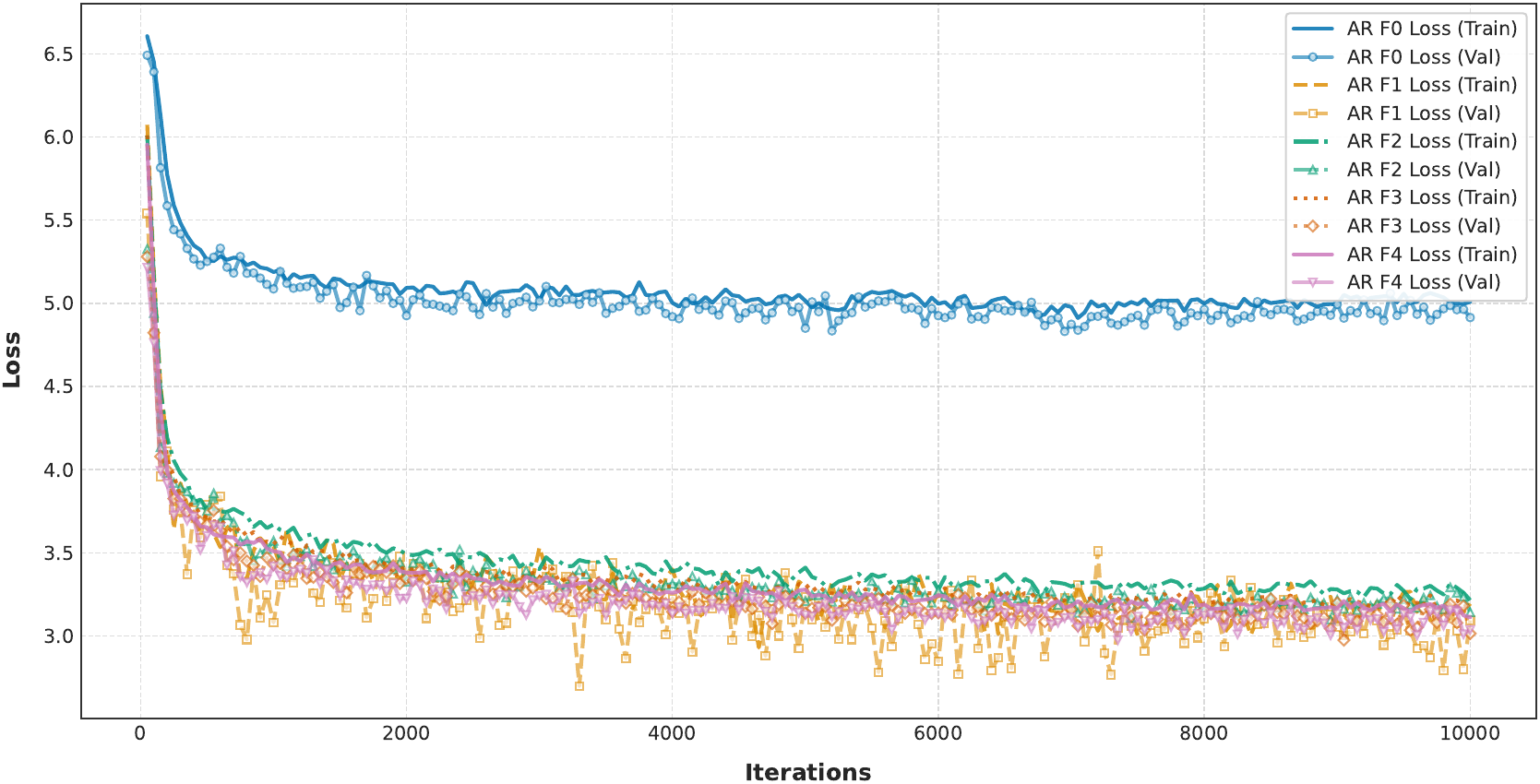}
    \centerline{\small (b) Autoregressive losses}
  \end{minipage}
  \caption{Training loss curves (3/3): detailed loss breakdown and autoregressive loss components.}
  \label{fig:loss_row3}
\end{figure}

We train FluentAvatar on a mixed dataset combining the super-resolved CMLR (Chinese) and the original HDTF (English) datasets for 10,000 steps. Figures~\ref{fig:loss_row1}--\ref{fig:loss_row3} illustrate the progression of various loss components during training, demonstrating stable convergence and the contribution of each component to the total loss.

\section{Ablation Study Details}
\label{appendix:ablation_details}

\subsection{Attention Mechanisms}
\label{appendix:attention_details}

We compare four attention configurations that differ in the scope of phoneme information accessible during frame generation:

\textbf{(1) Non-Causal Full Attention.} The model accesses the entire phoneme sequence (past and future) when generating any frame. This serves as a theoretical upper bound but is unsuitable for streaming generation.

\textbf{(2) Causal Accumulative Attention.} When generating the $i$-th frame, the model accesses all phonemes from the 1st to the $i$-th, representing standard autoregressive attention.

\textbf{(3) Limited History Attention.} A sliding window of size 2: the model accesses only the $i$-th and $(i{-}1)$-th phonemes, providing limited local context.

\textbf{(4) One-to-One Attention (Ours).} The model accesses only the corresponding $i$-th phoneme, enforcing the strictest phoneme-to-frame alignment without historical or future information.

\subsection{Frame Allocation Strategies}
\label{app:frame_allocation}

We compare three strategies with the keyframe generator (FKG) fixed:

\textbf{(1) Random.} The number of in-between frames per keyframe pair is randomly sampled, constrained to match the target video length.

\textbf{(2) Fixed.} A constant number of in-between frames is inserted for every pair, ignoring actual temporal intervals.

\textbf{(3) Dynamic (Ours).} Frames are allocated proportionally to the temporal interval between keyframes, enforcing a globally consistent frame rate.

As shown in Table~\ref{tab:frame_allocation}, the Dynamic strategy achieves the best FVD (503.29) and substantially lower Sync-D (10.90), confirming that timestamp-aware allocation improves temporal coherence and lip synchronization without sacrificing reconstruction quality.

\begin{table}[H]
  \centering
  \caption{Ablation of frame allocation strategies on CMLR.}
  \label{tab:frame_allocation}
  \scriptsize
  \setlength{\tabcolsep}{2pt}
  \renewcommand{\arraystretch}{1.0}
  \begin{tabular}{lcccccccc}
    \toprule
    Method & FID$\downarrow$ & FVD$\downarrow$ & LPIPS$\downarrow$ & PSNR$\uparrow$ & SSIM$\uparrow$ & CSIM$\uparrow$ & Sync-C$\uparrow$ & Sync-D$\downarrow$ \\
    \midrule
    Random & 24.82 & 553.39 & \textbf{0.40} & \textbf{15.07} & 0.49 & \textbf{0.85} & \textbf{1.09} & 14.11 \\
    Fixed & \textbf{24.56} & 522.32 & 0.41 & 15.04 & 0.49 & 0.84 & 0.94 & 14.08 \\
    Dynamic & 25.19 & \textbf{503.29} & \textbf{0.40} & \textbf{15.07} & 0.49 & \textbf{0.85} & 0.94 & \textbf{10.90} \\
    \bottomrule
  \end{tabular}
\end{table}

\subsection{Loss Function}
\label{appendix:loss_ablation}

We ablate the four loss components by incrementally adding terms to the CE-only baseline (Table~\ref{tab:appendix_loss_ablation}). Each component individually improves over the baseline, and their combination yields the best overall quality. Removing either identity or facial similarity loss leads to marked degradation, underscoring their importance in preserving identity and visual realism.

\begin{table*}[t]
\centering
\caption{Ablation of loss function combinations. CE: Cross-Entropy; LPIPS: Perceptual; Id.: Identity Consistency; Facial: Facial Similarity. Bold = best; underlined = second best.}
\label{tab:appendix_loss_ablation}
\small
\setlength{\tabcolsep}{6pt}
\renewcommand{\arraystretch}{1.1}
\begin{tabular}{lccccccccc}
  \toprule
  \multirow{2}{*}{\textbf{Config.}} & \multirow{2}{*}{\textbf{Exp.}} & \multicolumn{4}{c}{\textbf{Loss Functions}} & \multicolumn{4}{c}{\textbf{Evaluation Metrics}} \\
  \cmidrule(lr){3-6} \cmidrule(lr){7-10}
  & & CE & LPIPS & Id. & Facial & FID$\downarrow$ & LPIPS$\downarrow$ & PSNR$\uparrow$ & SSIM$\uparrow$ \\
  \midrule
  Baseline & 1 & \checkmark & & & & 28.14 & 0.0365 & 25.08 & 0.884 \\
  \midrule
  \multirow{3}{*}{+Single}
  & 2 & \checkmark & \checkmark & & & 16.65 & \textbf{0.0128} & \underline{32.57} & 0.962 \\
  & 5 & \checkmark & & \checkmark & & 16.20 & 0.0138 & 32.05 & 0.962 \\
  & 6 & \checkmark & & & \checkmark & 16.47 & \underline{0.0131} & 32.47 & \underline{0.965} \\
  \midrule
  \multirow{3}{*}{+Double}
  & 3 & \checkmark & \checkmark & \checkmark & & 15.66 & 0.0151 & 31.91 & 0.963 \\
  & 4 & \checkmark & \checkmark & & \checkmark & 18.14 & 0.0162 & 31.84 & 0.962 \\
  & 7 & \checkmark & & \checkmark & \checkmark & \textbf{13.44} & 0.0133 & 32.44 & 0.964 \\
  \midrule
  Full & 8 & \checkmark & \checkmark & \checkmark & \checkmark & \underline{13.86} & 0.0136 & \textbf{33.13} & \textbf{0.967} \\
  \bottomrule
\end{tabular}
\end{table*}

\subsection{Phoneme-level Conditioning}
\label{appendix:prl}

\begin{figure}[t]
  \centering
  \setlength{\abovecaptionskip}{4pt}
  \setlength{\belowcaptionskip}{1pt}
  \includegraphics[width=\columnwidth,keepaspectratio]{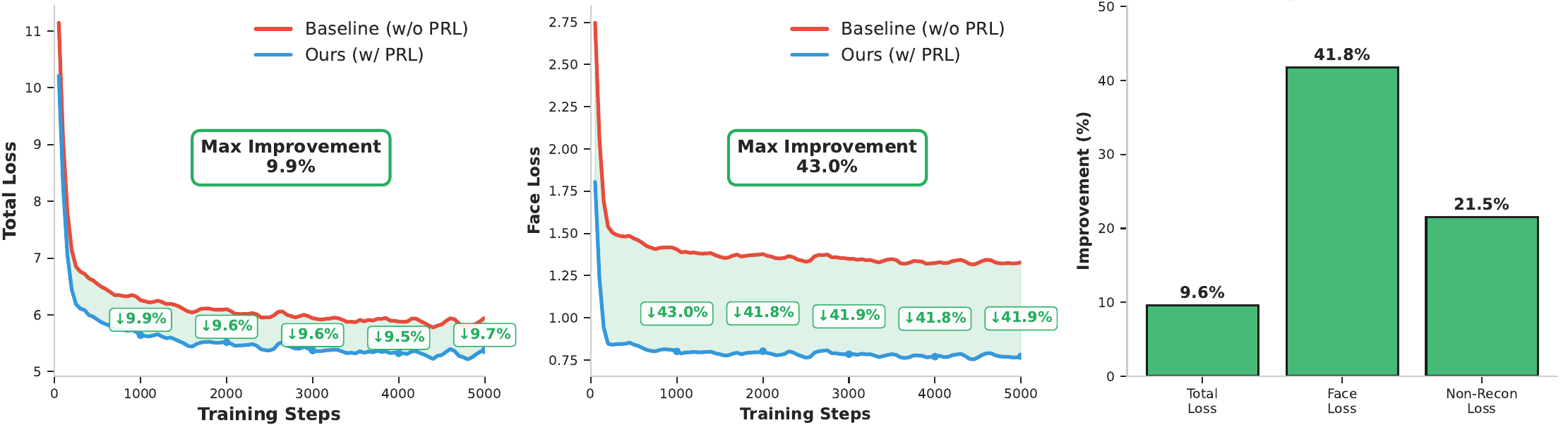}
  \caption{Training loss comparison: phoneme-level vs.\ word-level conditioning. Phoneme-level conditioning consistently lowers all loss components, with face reconstruction loss reduced by 41.8\%.}
  \label{fig:PVM_table}
\end{figure}

To assess whether phoneme-level conditioning is necessary, we compare it against word-level conditioning on CMLR. In the word-level baseline, keyframe generation is conditioned on audio features aligned with word-level units in the transcript; in the phoneme-level configuration (ours), the same audio is aligned at the phoneme level. As shown in Figure~\ref{fig:PVM_table}, phoneme-level conditioning reduces face reconstruction, total, and non-reconstruction training losses by 41.8\%, 9.6\%, and 21.5\% respectively, indicating faster convergence and more stable optimization. The substantially lower face reconstruction loss is particularly significant, as it directly reflects the model's ability to generate accurate facial details, confirming that finer-grained phoneme units provide more precise speech-lip alignment than coarser word-level units.

\section{Human Evaluation Details}
\label{app:human_eval}

We conducted a user study with 30 participants (15 with computer vision background, 15 general users) comparing FluentAvatar with eight state-of-the-art methods. Participants rated 5 video clips per method (45 clips total, selected to cover different speakers, both languages, and varying speech tempos) across four dimensions: Flicker, Temporal Coherence, Body Movement Realism, and Lip Synchronization, on a 5-point Likert scale. Videos were shuffled and anonymized to prevent ordering and naming bias. FluentAvatar achieves the highest average rating on all four dimensions ($4.8 \pm 0.41$, $4.4 \pm 0.50$, $3.8 \pm 0.48$, $3.6 \pm 0.62$), consistently outperforming competing approaches with low variance across participants.

\section{Inference Efficiency Analysis}
\label{app:inf_efficiency}

\begin{figure}[t]
  \centering
  \includegraphics[width=0.9\columnwidth]{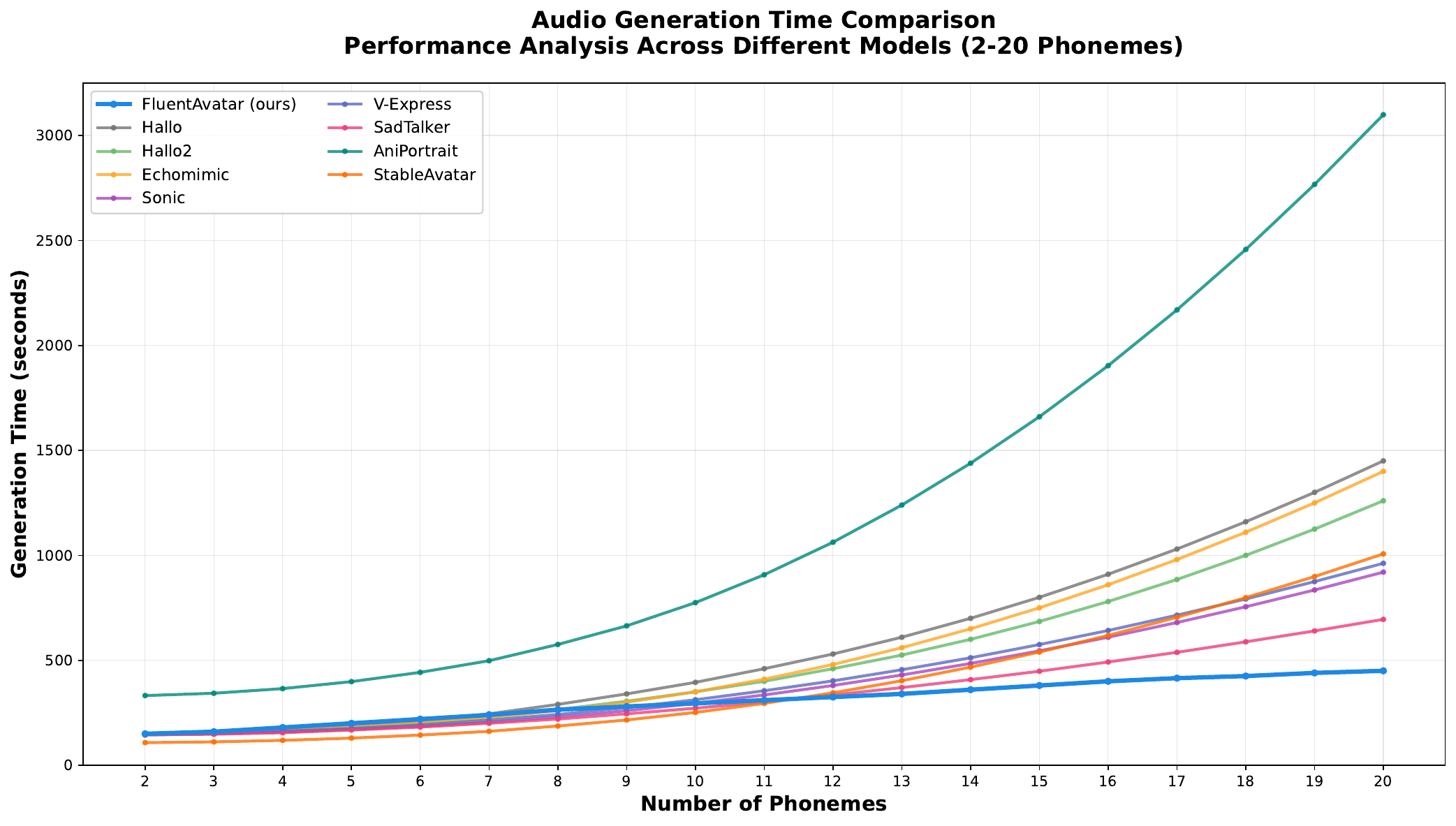}
  \caption{Generation time comparison. FluentAvatar scales nearly linearly with phoneme count.}
  \label{fig:times}
\end{figure}

Figure~\ref{fig:times} visualizes how inference time scales with input length. We vary the phoneme count from 2 to 20 and measure generation time for $512\times512$ videos. FluentAvatar exhibits near-linear growth, whereas diffusion baselines scale super-linearly. At 20 phonemes, FluentAvatar is approximately $2.4\times$ faster than Hallo, consistent with the parallel interpolation design.

\end{document}